\documentclass[10pt,journal,compsoc]{IEEEtran}

\ifCLASSOPTIONcompsoc
  \usepackage[nocompress]{cite}
\else
  \usepackage{cite}
\fi

\ifCLASSINFOpdf
\else
\fi

\usepackage{algorithmic}

\usepackage{subfigure}
\usepackage{multirow}
\usepackage{algorithm}
\usepackage{amsmath}
\usepackage{amsfonts}
\usepackage{graphicx}
\usepackage{makecell}
\usepackage{balance}
\usepackage{xcolor}

\usepackage{adjustbox}

\usepackage{booktabs}

\newtheorem{definition}{Definition}
\newcommand{\HL}{}
\newcommand{\yg}{}

\newcommand{\tabincell}[2]{\begin{tabular}{@{}#1@{}}#2\end{tabular}}  

\begin{document}

\title{Anomaly Detection in Dynamic Graphs via Transformer}

\author{Yixin Liu, Shirui Pan, Yu Guang Wang, Fei Xiong, Liang Wang, Qingfeng Chen,
        Vincent CS Lee
\IEEEcompsocitemizethanks{\IEEEcompsocthanksitem Y. Liu, S. Pan, and V. Lee are with the Department of Data Science and AI, Faculty of IT, Monash University, Clayton, VIC 3800, Australia\protect\\
E-mail: yixin.liu@monash.edu; shirui.pan@monash.edu; Vincent.CS.Lee@monash.edu
\IEEEcompsocthanksitem Y. G. Wang is with Shanghai Jiao Tong University, in Institute of Natural Sciences and School of Mathematical Sciences, and with the Max Planck Institute for Mathematics in Sciences, in Mathematics Machine Learning group Email:  yuguang.wang@sjtu.edu.cn
\IEEEcompsocthanksitem F. Xiong is with Key Laboratory of Communication and Information Systems, Beijing Municipal Commission of Education, Beijing Jiaotong University, Beijing 100044, China; Email:  xiongf@bjtu.edu.cn
\IEEEcompsocthanksitem L. Wang is with School of Computer Science, Northwestern Polytechnical University, Xi’an 10072, China; Email: liangwang@nwpu.edu.cn
\IEEEcompsocthanksitem Q. Chen is with School of Computer, Electronic and Information, Guangxi University, Nanning, 530004, China; Email: qingfeng@gxu.edu.cn
\IEEEcompsocthanksitem Corresponding Author: Shirui Pan}
}

\markboth{Journal of \LaTeX\ Class Files,~Vol.~14, No.~8, August~2015}%
{Shell \MakeLowercase{\textit{et al.}}: Bare Demo of IEEEtran.cls for Computer Society Journals}

\IEEEtitleabstractindextext{%
\begin{abstract}
Detecting anomalies for dynamic graphs has drawn increasing attention due to their wide \HL{applications} in social networks, e-commerce, and cybersecurity.  Recent deep learning-based approaches have shown promising results over shallow methods. However, they fail to address two core challenges of anomaly detection in dynamic graphs: the lack of informative encoding for unattributed nodes and the difficulty of learning discriminate knowledge from coupled spatial-temporal dynamic graphs. To overcome these challenges, in this paper, we present a novel \underline{\textbf{T}}ransformer-based \underline{\textbf{A}}nomaly \underline{\textbf{D}}etection framework for \underline{\textbf{DY}}namic graphs (\textbf{TADDY}). Our framework constructs a comprehensive node encoding strategy to better represent each node's structural and temporal roles in an evolving graphs stream. Meanwhile, TADDY captures informative representation from dynamic graphs with coupled spatial-temporal patterns via a dynamic graph transformer model. The extensive experimental results demonstrate that our proposed TADDY framework outperforms the state-of-the-art methods by a large margin on \HL{six} real-world datasets.
\end{abstract}

}

\maketitle

\IEEEdisplaynontitleabstractindextext

%
\IEEEpeerreviewmaketitle


\IEEEraisesectionheading{\section{Introduction}\label{sec:introduction}}

\IEEEPARstart{I}{n} \HL{recent years}, graphs have attracted a surge of research attention \HL{with the development of networked applications} in social networks \cite{wang2019influence}, human knowledge networks \cite{ji2020survey}, business networks \cite{zheng2020clustering} and cybersecurity \cite{gao2020hincti}.
However, the bulk of the existing researches \HL{focus} on static graphs \cite{jin2021survey,xia2021graph}, yet the real-world graph data often evolves over time \cite{peng2021lime,jiao2021temporal}. Taking social networks as an example, there are always fresh persons who enroll in the community every month, and the relation between individuals is changing over time. To model and analyze graphs where nodes and edges change over time, mining dynamic graphs gains increasing popularity in the community of graph analysis.

Among various analysis problems for dynamic graphs, detecting the anomalous edges in an evolving graph stream is a critical task \cite{yu2018netwalk,zheng2019addgraph}. Considering a user-item network in the e-commerce \yg{scenario}, the attackers tend to make fake purchase orders to increase the influence of certain goods illegally. It is of great significance to detect such fake orders \HL{to maintain} a fair trading environment.

\begin{figure}[htbp]
	\centering
	\includegraphics[width=0.5\textwidth]{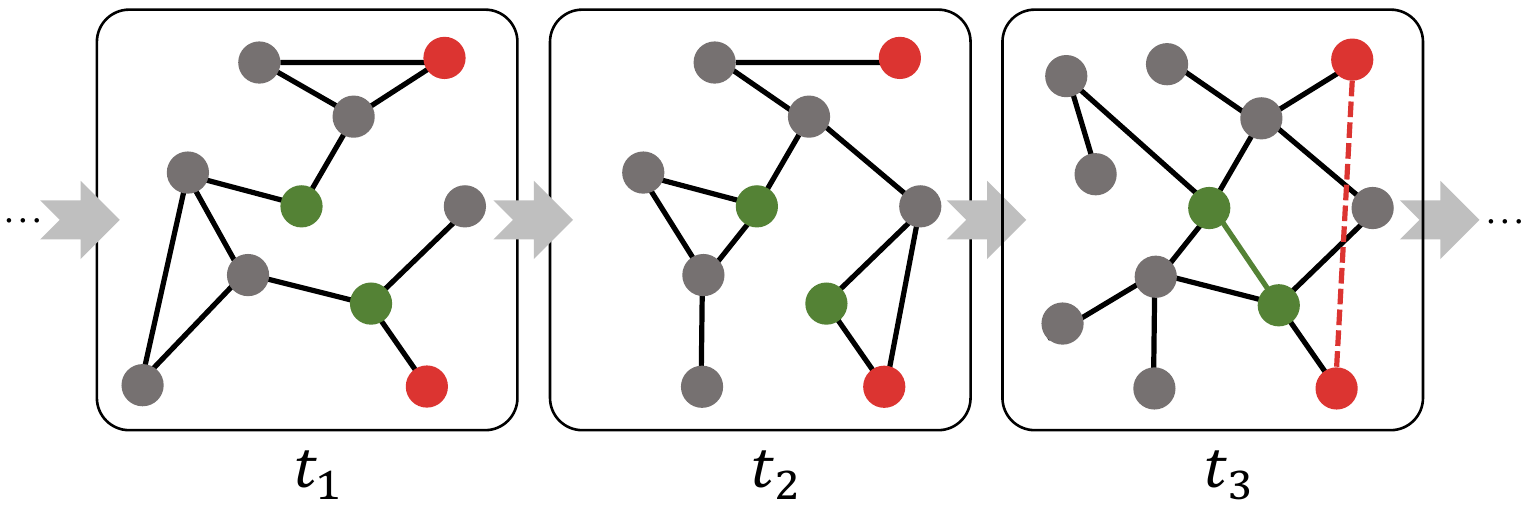}
	\caption{A toy example to illustrate how the coupled information affects the detection of edges' legality. The three graphs are a fragment from a dynamic graph stream at a sequential timeline $\{t_1,t_2,t_3\}$. The solid green line represents a normal edge at $t_3$, while the red dash line indicates an anomalous edge. We highlight the corresponding nodes in the previous timestamps with colors.}
	\vspace{-0.2cm}
	\label{fig:toy_example}
\end{figure}

Detecting anomalies in dynamic graphs, however, is not a trivial task since there are two challenges in dynamic graph learning. 
\textit{Challenge 1} is the lack of raw attribute information in most dynamic graphs. Due to the explosive demand for data volume of time-evolving attributes or the inaccessible attributes caused by privacy issues, it is hard to construct attribute information to represent each node from the mainstream raw dynamic graph datasets. 
\HL{To fill the gap}, an effective encoding method \HL{that constructs artificial features to represent evolving nodes} is required. 
\textit{Challenge 2} is the difficulty of learning discriminative knowledge from dynamic graphs where spatial (structural) information and temporal information are coupled. Figure \ref{fig:toy_example} provides a toy example to illustrate how coupled information affects the detection of edge abnormality. The green edge tends to be normal since there are close structural communications between their neighborhoods in the previous timestamps. The red edge, on the contrary, is an anomalous edge with a high probability because the two red nodes always keep a distance from each other in the former snapshots. The point is that both structural \HL{(i.e., shared neighborhoods)} and temporal \HL{(i.e., previous interaction)} factors should be considered simultaneously when making decisions, \HL{raising the} challenge in understanding such coupled information.

Aiming to detect anomalies in dynamic graphs, various types of approaches are proposed in the recent decade. The shallow methods like GOutlier \cite{aggarwal2011outlier} and CM-Sketch \cite{ranshous2016scalable} utilize shallow learning mechanisms (\HL{e.g.}, structural connectivity model or historical behavior analysis) to detect anomalies. However, empirical experiments \HL{show that} these methods \HL{suffer from} limited performance when detecting anomalous edges in large and complex dynamic graphs \cite{yu2018netwalk}. Very recently, as a novel branch, deep learning-based methods\HL{,} have shown to be a powerful solution for dynamic graph learning. For example, NetWalk \cite{yu2018netwalk} leverages dynamic deep graph embedding technique with a clustering-based detector to detect anomalies; AddGraph\cite{zheng2019addgraph}, StrGNN \cite{cai2020structural} and H-VGRAE \cite{yang2020h} further exploit end-to-end deep neural network models to solve the problem.

Despite their improved performance, the existing deep learning-based methods fail to address the \HL{aforementioned} challenges very well. 
\HL{Specifically}, when facing the lack of raw node attributes, they do not create informative node \HL{encodings} to represent the nodes' \HL{properties}. The one-hot identity features in \cite{yu2018netwalk,yang2020h} and the random initialized features in \cite{zheng2019addgraph} cannot express any structural or temporal property of each node. The distance-based node labeling strategy in \cite{cai2020structural} only considers local structural information, \HL{limiting its} \yg{expressive power}.
Furthermore, most of \HL{them} use two individual network modules to extract spatial and temporal features, resulting in their insufficient capability to capture the coupled information. For instance, in AddGraph \cite{zheng2019addgraph} and StrGNN \cite{cai2020structural}, Graph Convolutional Networks (GCNs) are employed to acquire spatial knowledge, with following Gated Recurrent Units (GRUs) \HL{capturing} temporal information. The isolated processing of two types of information results in missing the coupled spatial-temporal features and further leads to a sub-optimal solution.

Aiming to resolve these challenges, in this paper, we propose a novel \underline{\textbf{T}}ransformer-based \underline{\textbf{A}}nomaly \underline{\textbf{D}}etection framework for \underline{\textbf{DY}}namic graph (\textbf{TADDY} for abbreviation). Our theme is to construct a node encoding to cover sufficient spatial and temporal knowledge and leverage a sole transformer model to capture \HL{the} coupled spatial-temporal information. 
More specifically, to overcome \textit{Challenge 1}, we carefully design a comprehensive node encoding composed of three functional terms \yg{to distill} global spatial, local spatial, and temporal information. Learnable mapping functions are integrated into the node encoding, which helps the framework automatically extract informative encoding in an end-to-end manner. For \textit{Challenge 2}, we develop a dynamic graph transformer model to simultaneously learn spatial and temporal knowledge. An edge-based substructure sampling is performed to capture contextual information crossing time as the input of the transformer model. Then, the coupled spatial-temporal information is extracted by the attention mechanism crossing structure and time. To sum up, the main contributions of this paper are:

\begin{itemize} 
	\item We propose an end-to-end transformer-based learning framework, TADDY, for anomaly detection on dynamic graphs. This is the first transformer-based method for dynamic graph learning and graph anomaly detection to the best of our knowledge.
	
	\item We design a comprehensive encoding method for nodes in dynamic graphs. The proposed node encoding integrates various knowledge, including global spatial, local spatial and temporal information.
	
	\item We present a dynamic graph transformer model which aggregates spatial and temporal knowledge simultaneously. A novel edge-based substructure sampling strategy is leveraged to provide sufficient receipt fields for the learning model.
	
	\item We evaluate the effectiveness of TADDY on \HL{six} benchmark datasets. The extensive experiments demonstrate that our method \yg{delivers state-of-the-art performance}.
\end{itemize}

We organize the rest of this paper as follows. The related works are reviewed in Section \ref{sec:rw}. We describe the problem definition in Section \ref{sec:definition}. In Section \ref{sec:method} the overall pipeline and each component of our framework are introduced. The experimental results are demonstrated in Section \ref{sec:exp}. Finally, we conclude the contributions and future works of this work in Section \ref{sec:conclusion}.

\section{Related Work} \label{sec:rw}

This section briefly reviews \yg{existing anomaly detection methods} for dynamic graphs and transformers.

\subsection{Anomaly Detection in Dynamic Graphs} \label{subsec:rw_ad}

Anomaly detection in dynamic graphs attracts considerable interest by the research community \cite{peng2021streaming}, for which, many methods \yg{have been proposed in recent years.}
For example, GOutlier \cite{aggarwal2011outlier} employs a structural connectivity model to detect outliers in graph streams and builds dynamic network partition to maintain the connectivity behavior model. 
CAD \cite{sricharan2014localizing} detects node relationships by tracking a measure that combines information regarding changes in graph structure \yg{and in edge} weights. 
CM-Sketch \cite{ranshous2016scalable} considers both the local structural information and historical behavior to discriminate the edge's anomalous property. 
StreamSpot \cite{manzoor2016fast} is a clustering-based approach that utilizes a novel similarity function for heterogeneous graphs property comparison and leverages a centroid-based clustering method to model the behaviors of graph stream. 
SpotLight \cite{eswaran2018spotlight} uses a randomized sketching technique to guarantee a large mapped distance between anomalous and normal instances in the sketch space. 
Since these approaches leverage the shallow mechanisms to detect the anomalous edges, we categorize them into shallow learning-based methods. 

Another branch of approach employs deep learning technique 
to capture anomalous data in dynamic graphs, which is denoted as the category of deep learning-based method. 
NetWalk \cite{yu2018netwalk} leverages a random walk-based encoder to generate node embeddings with clique embedding objective and then models the network evolving via dynamic updating reservoirs. Finally, a dynamic clustering-based anomaly detector is employed to score the abnormality of each edge.
AddGraph \cite{zheng2019addgraph} further constructs an end-to-end neural network model to capture dynamic graphs' spatial and temporal patterns. A GCN \cite{kipf2017semi} is served as a structural features extractor, and a GRU-attention module is designed to combine short-term and long-term dynamic evolving.
StrGNN \cite{cai2020structural} extracts the $h$-hop enclosing subgraph of edges and leverages stacked GCN \cite{kipf2017semi} and GRU to capture the spatial and temporal information. The learning model is trained in an end-to-end way with negative sampling from “context-dependent” noise distribution.
H-VGRAE \cite{yang2020h} builds a hierarchical model by combining variational graph autoencoder and recurrent neural network. To detect anomalous edges, the edge reconstruction probability is used to measure the abnormality.

Our proposed TADDY framework can be categorized into the deep learning-based methods but has two main differences compared to the \yg{existing} approaches mentioned above. 
\yg{Most of} the above approaches employ different network modules to separately extract spatial and temporal features, while TADDY uses a transformer network to model spatial and temporal information simultaneously.

Secondly, these methods consider naive node encoding from unattributed dynamic graphs as the network input, which may fail to provide sufficient information for the downstream neural network. 
In contrast to them, TADDY constructs a comprehensive node encoding that includes both spatial and temporal information.

\subsection{Transformers} \label{subsec:rw_transformer}

Transformers are a family of neural networks solely based on attention mechanisms to learn representative embedding for various data. 
The Transformer model is first proposed in \cite{vaswani2017attention}, which focuses on the machine translation tasks in natural language processing (NLP). 
BERT \cite{devlin2018bert} further applies transformers to multiple deep language understanding tasks by introducing the pre-training technique. 
Following Transformer and BERT, a large number of variant works are presented and reach state-of-the-art results on various NLP tasks \cite{lan2019albert,liu2019roberta,yang2019xlnet}.
Very recently, the transformer model is extended to the field of computer vision (CV) \cite{liu2021a}. 
For instance, DETR \cite{carion2020end} first leverages transformers on the object detection task. 
ViT \cite{dosovitskiy2021an} splits an image into multiple patches and uses a pure transformer model to learn the representation for image classification directly. 
SETR \cite{zheng2020rethinking} utilizes a ViT-like encoder for feature extraction and adopts a multi-level feature aggregation module for pixel-wise image segmentation. 
For further details about transformers on NLP and CV please see related surveys \cite{han2020survey,tay2020efficient}.

Some recent works also introduce transformers to the field of graph machine learning. 
GTN \cite{yun2019graph} is performed on heterogeneous graphs with transformers by meta-path-based relation learning. 
HGT \cite{hu2020heterogeneous} is a transformer model for the representation learning on \yg{web-scale} heterogeneous graphs, which reaches state-of-the-art results on various downstream tasks. 
GROVER \cite{rong2020self} integrates the message passing mechanism into the transformer architecture to learn representation for molecule graph data. 
Graph-BERT \cite{zhang2020graph} constructs a BERT-like network model for static graph learning and introduces various well-designed tasks for self-supervised model pre-training \cite{liu2021graph}.

Our proposed framework introduces transformers as our backbone neural network model due to its powerful \yg{expressive capability}. Differently, we extend transformers to dynamic graphs, which is a more complex learning scenario where both structural and temporal features should be considered. By comparison, most of the existing methods focus on static graphs.

\begin{table}[t]
\centering
\caption{\HL{Commonly used notation with explanations.}} 
\vspace{-0.1cm}
\begin{adjustbox}{width=1\columnwidth,center}
\begin{tabular}{ l|l}  
\toprule[1.0pt]
Notation & Explanation  \\
\cmidrule{1-2}
$\mathbb{G}=\{\mathcal{G}^t\}^{T}_{t=1}$ & A graph steam with a maximum timestamp of $T$. \\
$\mathcal{G}^t=(\mathcal{V}^t,\mathcal{E}^t)$ & The snapshot graph at timestamp $t$. \\
$\mathcal{V}^t$ & The node set at timestamp $t$. \\
$\mathcal{E}^t$ & The edge set at timestamp $t$. \\
$v^t_{i}  \in \mathcal{V}^t$ & A node with index $i$ at the timestamp $t$. \\ 
$e^t_{i,j} = (v_i^t, v_j^t) \in \mathcal{E}^t$ & An edge between $v_i^t$ and $v_j^t$ at the timestamp $t$. \\ 
$n^t$ & The number of nodes at timestamp $t$. \\
$m^t$ & The number of edges at timestamp $t$. \\
$\mathbf{A}^t$ & The binary adjacency matrix at timestamp $t$. \\
$f(\cdot)$ & Anomaly score function. \\
\cmidrule{1-2}
\tabincell{l}{$\mathbb{G}^t_\tau = $ \\ $\{\mathcal{G}^{t-\tau+1}, \cdots, \mathcal{G}^{t} \}$} & \tabincell{l}{The sequence of graphs with timestamp $t$ as the end and \\ $\tau$ as the window size (length).} \\
$\mathcal{S}(e_{\rm tgt}^t)$ & The substructure node set of target edge $e_{\rm tgt}^t$. \\
\cmidrule{1-2}
$\mathbf{x}_{\rm diff}(v_j^i)$ & The diffusion-based spatial encoding of node $v_j^i$. \\
$\mathbf{x}_{\rm dist}(v_j^i)$ & The distance-based spatial encoding of node $v_j^i$. \\
$\mathbf{x}_{\rm temp}(v_j^i)$ & The relative temporal encoding of node $v_j^i$. \\
$\mathbf{x}(v_j^i)$ & The fused encoding of node $v_j^i$. \\
$\mathbf{X}(e^t_{\rm tgt})$ & The encoding matrix of target edge $e_{\rm tgt}^t$. \\
\cmidrule{1-2}
$\mathbf{H}^{(l)}$ & The output embedding of the $l$-th layers of Transformer. \\
$\mathbf{Q}^{(l)}$ & The query matrix of the $l$-th layers of Transformer. \\
$\mathbf{K}^{(l)}$ & The key matrix of the $l$-th layers of Transformer. \\
$\mathbf{V}^{(l)}$ & The value matrix of the $l$-th layers of Transformer. \\
$\mathbf{z}(e_{\rm tgt}^t)$ & The embedding of target edge $e_{\rm tgt}^t$. \\
$\mathbf{W}_{Q}^{(l)}, \mathbf{W}_{K}^{(l)}, \mathbf{W}_{V}^{(l)}$ & The learnable parameters of Transformer. \\
\cmidrule{1-2}
$e_{{\rm pos},i} \in \mathcal{E}^t$ & The $i$-th positive edge from $\mathcal{E}^t$. \\
$e_{{\rm neg},,i} \in \mathcal{E}^t_n \sim P_{n}(\mathcal{E}^t)$ & The $i$-th negative edge by negative sampling $P_{n}(\mathcal{E}^t)$. \\
$\mathbf{w}_S, b_S$ & The learnable parameters of Anomaly Detector. \\
\cmidrule{1-2}
$k$ & The number of contextual nodes. \\
$\tau$ & The size of time window. \\
$d_{enc}$ & The dimension of encoding. \\
$d_{emb}$ & The dimension of embedding. \\
$L$ & The number of layers of Transformer. \\
\bottomrule[1.0pt]
\end{tabular}
\end{adjustbox}
\vspace{-0.2cm}
\label{table:notation}
\end{table}

\section{Problem Definition} \label{sec:definition}

In this paper, we model a dynamic graph as a graph stream represented by a series of discrete snapshots. The definition of dynamic graphs is given as follows:

\begin{definition}
	 Considering a dynamic graph with a maximum timestamp of $T$, a graph steam is represented by $\mathbb{G}=\{\mathcal{G}^t\}^{T}_{t=1}$, where each $\mathcal{G}^t=(\mathcal{V}^t,\mathcal{E}^t)$ is the snapshot at timestamp $t$, $\mathcal{V}^t$ is the node set \HL{at} timestamp $t$, and $\mathcal{E}^t$ is the edge set \HL{at} timestamp $t$. An edge $e^t_{i,j} = (v_i^t, v_j^t) \in \mathcal{E}^t$ indicates that there is a connection between node $v_i^t$ and $v_j^t$ at the timestamp $t$, where $v_i^t, v_j^t \in \mathcal{V}^t$. We use $n^t=\left| \mathcal{V}^t \right|$ and $m^t=\left| \mathcal{E}^t \right|$ to denote the number of nodes and edges \HL{at} timestamp $t$ respectively. A binary adjacency matrix $\mathbf{A}^t \in \mathbb{R}^{n^t \times n^t}$ is employed to denote $\mathcal{G}^t$, where $\mathbf{A}^t_{i,j} = 1$ if there is a link between nodes $v_i$ and $v_j$ at timestamp $t$, otherwise $\mathbf{A}^t_{i,j} = 0$.
\end{definition}

The goal of this paper is to detect the anomalous edges in each timestamp. According to the aforementioned \yg{notation}, we formalize the anomaly detection in dynamic graphs as a scoring problem:

\begin{definition}{\textbf{Anomaly detection in dynamic graphs.}}
	Given a dynamic graph $\mathbb{G}=\{\mathcal{G}^t\}^{T}_{t=1}$ where each $\mathcal{G}^t=(\mathcal{V}^t,\mathcal{E}^t)$, for each $e^t_{i,j} \in \mathcal{E}^t$, the goal of anomaly detection is to produce the anomaly score $f(e^t_{i,j})$, where $f(\cdot)$ is a learnable anomaly score function. The anomaly score indicates the abnormality degree of the edge, where a larger score $f(e^t_{i,j})$ shows a higher anomalous probability of $e^t_{i,j}$. 
\end{definition}

Following the previous works \cite{yu2018netwalk,zheng2019addgraph,cai2020structural}, we consider an unsupervised setting for anomaly detection in dynamic graphs. Specifically, in the training phase, no labeled data for \HL{anomalies} is given, but we assume that all edges in the training set are normal. The binary labels of abnormality are given in the testing phase to evaluate the performance of algorithms. Concretely, a label $y_{e^t_{i,j}} = 1$ indicates that edge $e^t_{i,j}$ is an anomalous edge, and $y_{e^t_{i,j}} = 0$ indicates that $e^t_{i,j}$ is normal. Note that the \HL{distribution of normal and anomalous edges} is often imbalanced, which means the number of normal edges is much larger than anomalous edges.

\HL{For the convenience of readers, the notation used in this paper is summarized in Table \ref{table:notation}.}

\section{Methodology} \label{sec:method}

\begin{figure*}[htbp]
	\centering
	\includegraphics[width=0.9\textwidth]{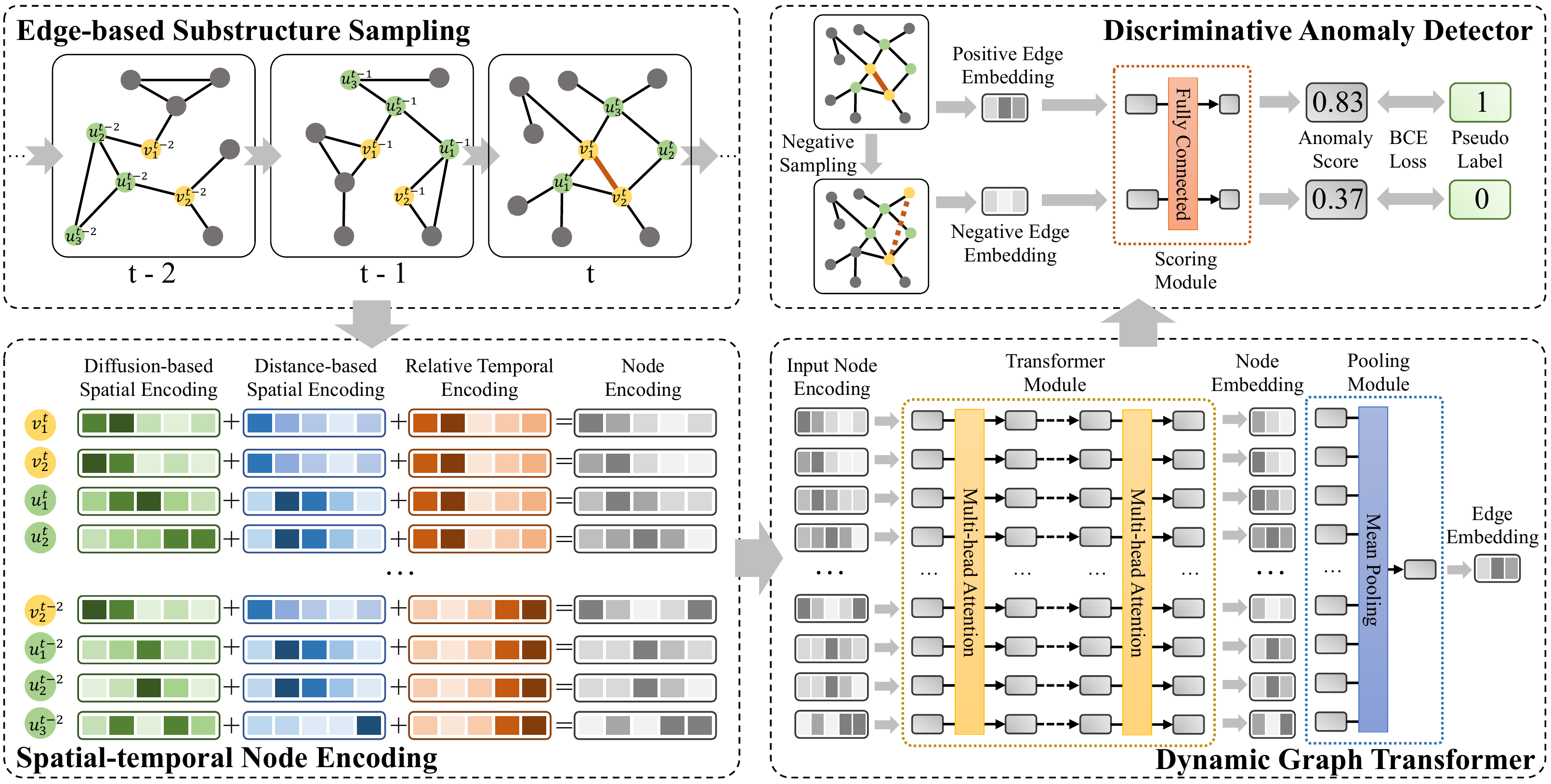}
	\caption{\HL{The} overall framework of TADDY. The framework is composed of four components: edge-based substructure sampling, spatial-temporal node encoding, dynamic graph transformer, and discriminative anomaly detector. Here we regard the edge in red $\{v^t_1,v^t_2\}$ as a target edge and exhibit a running example. In edge-based substructure sampling, the target nodes (in yellow) and contextual nodes (in green) from multiple timestamps are sampled to construct the substructure node set, where \HL{neighboring node number} $k$ and \HL{time window size} $\tau$ are both set to be $3$. Then, three types of encoding are computed for each node and further fused into the node encoding. Taking the node encoding as input data, the dynamic graph transformer learns latent node embedding with attention layers and leverages mean pooling to calculate the edge embedding. Finally, in the discriminative anomaly detector, the negative \HL{edges are} acquired by negative sampling. The scoring module computes the anomaly scores for positive and negative edges. The whole framework is trained with a binary cross-entropy loss in an end-to-end manner.}
	\label{fig:framework}
\end{figure*}

In this section, we introduce the general framework of TADDY. The overview of our proposed framework is illustrated in Figure \ref{fig:framework}. On the highest level, TADDY consists of four components, namely \textit{edge-based substructure sampling}, \textit{spatial-temporal node encoding}, \textit{dynamic graph transformer}, and \textit{discriminative anomaly detector}. The framework \HL{is trained} in an end-to-end manner, \HL{indicating} that the anomaly scores are output and learned directly.
At first, to capture the spatial-temporal contexts of each target edge, we perform \textit{edge-based substructure sampling} to acquire the target nodes and the contextual nodes in multiple timestamps.
Then, the \textit{spatial-temporal node encoding} generates the node encodings as the input of the transformer model. Both spatial and temporal information for each node are integrated into a fixed-length encoding.
After that, the \textit{dynamic graph transformer} extracts the spatial-temporal knowledge of edges via a sole transformer model composed of a transformer module and a pooling module.
Finally, in the \textit{discriminative anomaly detector}, we perform a negative sampling to generate pseudo negative edges, and an edge scoring module trained by binary cross-entropy loss is employed to calculate the output anomaly scores.
From Section \ref{subsec:sampling} to Section \ref{subsec:detector}, we concretely introduce the four main components of TADDY framework. In Section \ref{subsec:complexity}, we analyze the time complexity of our proposed framework.

\subsection{Edge-based Substructure Sampling} \label{subsec:sampling}

As is noticed in previous works \cite{cai2020structural,liu2021anomaly}, \HL{anomalies} often \HL{occur} in local \HL{substructures} of \HL{graphs}, indicating that we should zoom our \yg{receptive field} to a suitable local scale.
Therefore, instead of working on a complete dynamic graph, we first sample substructures as the data elements of our anomaly detection framework. Since we focus on detecting anomalous edges, we perform an edge-based sampling: each edge in dynamic graphs is viewed as the center of the sampled substructure and is denoted as a \textit{target edge}. For a given target edge, we denote the source node and destination node as \textit{target nodes}.

\HL{In addition to} the target nodes, it is necessary to include other neighboring nodes in the sampled substructure. In this paper, we denote these neighboring nodes as \textit{contextual nodes}. To acquire contextual information, a natural question arises here is: \textit{how to efficiently sample contextual nodes from a given target edge in a dynamic graph?} We first consider the sampling problem in a single \HL{snapshot} which can be regarded as a static graph. A naive solution is to extract the $h$-hop neighbors of target nodes as contexts. However, this strategy has several drawbacks. First, with $h$-hop neighbors sampling, the \HL{imbalanced} distribution of node degrees in real-world datasets would lead to a performance decline and low efficiency. For example, the average degree of UCI Messages dataset is $14.47$, while its maximum degree is $255$. 
\HL{For those popular nodes with high degrees, the numbers of $h$-hop neighbors would be explosive, resulting in the noisy information in sampled contexts and damaging the running efficiency.} 
Second, sampling $h$-hop neighbors ignores the different roles and importance of nodes in the substructure. Considering two target nodes with both shared and exclusive neighbors, it is obvious that the shared neighbors contribute more to detecting the target edges. However, \HL{this} simple strategy just views the shared and exclusive neighbors equally when sampling contextual nodes.

To address the aforementioned limitations, in this work, we borrow the graph diffusion technique \cite{klicpera2019diffusion,hassani2020contrastive} to sample a \HL{fixed-size} and importance-aware contextual node set for each target edge. With graph diffusion, a global view of graph structure is acquired, and then we can quantify the importance of each node for a given target node/edge. Formally, given an adjacency matrix of a static graph $\mathbf{A} \in \mathbb{R}^{n \times n}$, \yg{we define graph diffusion $\mathbf{S} \in \mathbb{R}^{n \times n}$ by}

\begin{equation}
\label{eq:diffusion}
\mathbf{S}=\sum_{k=0}^{\infty} \theta_{k} \mathbf{T}^{k},
\end{equation}

\noindent where $\mathbf{T} \in \mathbb{R}^{n \times n}$ is the generalized transition matrix and $\theta_{k}$ is the weighting coefficient which determines the ratio of global-local information. To guarantee convergence, some stricter conditions are considered which requires that $\sum_{k=0}^{\infty} \theta_{k}=1, \theta_{k} \in[0,1]$ and the eigenvalues $\lambda_i$ of $\mathbf{T}$ are bounded by $\lambda_i \in[0,1]$. By applying specific definitions of $\mathbf{T}$ and $\theta$, different instantiations of graph diffusion can be computed. For instance, Personalized PageRank (PPR) \cite{page1999pagerank} and the heat kernel \cite{kondor2002diffusion} are two popular examples of diffusion. Concretely, PPR chooses $\mathbf{T}=\mathbf{A} \mathbf{D}^{-1}$ and $\theta_{k}=\alpha(1-\alpha)^{k}$, where $\mathbf{D} \in \mathbb{R}^{n \times n}$ is the diagonal degree matrix and $\alpha \in (0,1)$ is the teleport probability. The heat kernel considers $\mathbf{T}=\mathbf{A} \mathbf{D}^{-1}$ and $\theta_{k}=e^{-\beta} \beta^{k} / k !$, \HL{where $\beta$ is the diffusion time}. To avoid multiple steps of iteration, the \yg{solutions to PPR and heat kernel can be} formulated as:

\begin{equation}
\label{eq:ppr}
\mathbf{S}^{\mathrm{PPR}}=\alpha\left(\mathbf{I}_{n}-(1-\alpha) \mathbf{D}^{-1 / 2} \mathbf{A} \mathbf{D}^{-1 / 2}\right)^{-1},
\end{equation}

\begin{equation}
\label{eq:heat}
\mathbf{S}^{\text {heat }}=\exp \left(\beta \mathbf{A} \mathbf{D}^{-1}- \beta \right).
\end{equation}

Given a diffusion matrix $\mathbf{S}$, a row $\mathbf{s_i}$ indicates the connectivity between the $i$-th node and each node from a global perspective. For example, $s_{i,j}$ represents the degree of connectivity between the $i$- and $j$- th nodes with a continuous value. By leveraging \HL{this} property, we can pick a fixed number of the most important enclosing nodes for a given target edge. Taking edge $e_{\rm tgt} = (v_1,v_2)$ as an example, we can compute the connectivity vector of $e_{\rm tgt}$ by adding the connectivity vectors of two target nodes:

\begin{equation}
\label{eq:edge_com}
\mathbf{s}_{e_{\rm tgt}}= \mathbf{s}_{v_1} + \mathbf{s}_{v_2}.
\end{equation}

\noindent Then, we sort the connectivity vector $\mathbf{s}_{e_{\rm tgt}}$ and select the top-$k$ nodes with the larger values to form the contextual node set $\mathcal{U}(e_{\rm tgt})$. Note that the target nodes themselves should be excluded when selecting the top-$k$ connectivity nodes. Finally, the sampled node set for substructure can be denoted as the union of contextual node set and target nodes, which can be formalized as $\mathcal{S}(e_{\rm tgt}) = \{v_1,v_2,\mathcal{U}(e_{\rm tgt})\}$.

According to the diffusion-based sampling, we can generate the contextual nodes from a \HL{single} static graph. However, for dynamic graphs, multiple timestamps should be considered to capture the dynamic evolving. \HL{Here}, we simply extend the static sampling method to dynamic graphs. \HL{Given} a target edge $e_{\rm tgt}^t = (v_1^t,v_2^t)$ at timestamp $t$, \HL{we consider a sequence of graphs $\mathbb{G}^t_\tau = \{\mathcal{G}^{t-\tau+1}, \cdots, \mathcal{G}^{t} \}$ with length $\tau$, where the time window size $\tau$ is a hyper-parameter and determines the receipt fields on time axis. With a sliding window mechanism, TADDY is able to capture dynamic evolving between timestamps $(t-\tau+1)$ and $t$. Then,} for each $\mathcal{G}^i \in \mathbb{G}^t_\tau$, we calculate the diffusion matrix $S^i$ and acquire the corresponding connectivity vector $\mathbf{s}^i_{e_{\rm tgt}^t}$. By picking the top-$k$ nodes and adding the target nodes, the substructure node set of the $i$-th timestamp can be sampled as $\mathcal{S}^i(e^t_{\rm tgt})$. By integrating the node set of multiple timestamps together, we can obtain the final substructure node set $\mathcal{S}(e_{\rm tgt}^t) = \bigcup_{i={t-\tau+1}}^{t}  \mathcal{S}^i(e_{\rm tgt}^t)$.

\subsection{Spatial-temporal Node Encoding} \label{subsec:encoding}

Unlike image and attributed graph data where each data instance (e.g., image patch or node) has its raw features, the dynamic graphs we study in this paper are often unattributed, which indicates that it is hard to find naturally appropriate data as the input of neural network models. This raises the question of \textit{how to construct an informative encoding as network input from unattributed dynamic graphs}. Similar to the concept of one-hot word encoding in NLP, an available solution is to use identity node encoding as the raw node feature, where a unique one-hot vector represents each node. However, identity node encoding has two limitations. First, the one-hot encoding is unable to contain enough structural and temporal information. The one-hot encoding only indicates the nodes' identity, but hard to express its structural roles and temporal status. Second, the identity node encoding is not friendly to large-scale and node-changing dynamic graphs. 
\HL{Third}, the fixed dimension cannot adapt to the dynamic changing set of nodes which is a common situation in dynamic graphs.

Inspired by the positional encoding in Transformer \cite{vaswani2017attention}, we introduce a novel spatial-temporal node encoding for dynamic graph transformers. The proposed node encoding consists of three components, namely diffusion-based spatial encoding, distance-based spatial encoding, and relative temporal encoding. The two terms of spatial encoding represent the structural role of each node from a global and local perspective respectively. The temporal encoding term, differently, provides the temporal information of each element in the substructure node set. To the end, the three encoding terms are fused \HL{as} the input node encoding which contains comprehensive spatial-temporal information. \HL{Note that we generate the encoding by learnable linear projections instead of frequency-aware sin/cos functions used in \cite{vaswani2017attention}. The reason is that the learnable functions are more flexible to model the correlations between different timestamps or positions.}

In the rest of this subsection, we discuss the three encoding terms sequentially and then introduce the encoding fusion operation.

\subsubsection{Diffusion-based Spatial Encoding} \label{subsec:diffusion_encoding}

As is introduced in Section \ref{subsec:sampling}, graph diffusion provides a global view \HL{of} the structural role of each node. With the edge connectivity vector $\mathbf{s}_{e_{\rm tgt}}$ computed by Eq. (\ref{eq:edge_com}), it is easy to acquired the strength of connectivity between the target edge and contextual node. Such \HL{a} property inspires us to design a spatial encoding that bases on graph diffusion. To prevent the indistinguishable encoding caused by similar diffusion values, we do not adopt the raw diffusion values directly but use a rank-based encoding. Specifically, for each node in a single-timestamp substructure node set $v_j^i \in \mathcal{S}^i(e^t_{\rm tgt})$, we sort nodes according to their diffusion values and adopt the ranking as the data source. According to the ranks, we compute the diffusion-based spatial encoding with a learnable encoding function (a single-layer linear mapping), which is similar to the learned positional encoding in \cite{gehring2017convolutional,vaswani2017attention}. The definition of the diffusion-based spatial encoding is given as:

\begin{equation}
\label{eq:diff_enc}
\mathbf{x}_{\rm diff}(v_j^i)= linear(rank(\mathbf{s}^i_{e_{\rm tgt}}[idx(v_j^i)])) \in \mathbb{R}^{d_{enc}},
\end{equation}

\noindent where $idx(\cdot)$, $rank(\cdot)$ and $linear(\cdot)$ are the index enquiring function, ranking function and learnable linear mapping respectively, and $d_{enc}$ is the dimension of node encoding. 

\subsubsection{Distance-based Spatial Encoding} \label{subsec:distance_encoding}

While the diffusion-based spatial encoding capture the global structural information, the local roles of each node should also be considered. Since the transformer model does not take the graph structure (e.g., adjacency matrix) as input like GNNs, we design a distance-based spatial encoding to represent the local connection around the target edge. Concretely, for each node in a single-timestamp substructure node set $v_j^i \in \mathcal{S}^i(e^t_{\rm tgt})$, we denote its distance to the target edge as the data source for encoding. The distance to the target edge can be decomposed into the minimum value of the relative distances to the two target nodes. For the target nodes themselves, the distances are denoted as $0$. A single-layer linear mapping is served as the learnable encoding function here, which is the same as the diffusion-based encoding. Formally, the distance-based spatial encoding can be expressed as:

\begin{equation}
\label{eq:dist_enc}
\mathbf{x}_{\rm dist}(v_j^i)= linear(min(dist(v_j^i,v_1^i),dist(v_j^i,v_2^i)) \in \mathbb{R}^{d_{enc}},
\end{equation}

\noindent where $dist(\cdot)$, $min(\cdot)$ and $linear(\cdot)$ are the relative distance computing function, minimum value function and learnable linear mapping respectively, and $d_{enc}$ is the dimension of node encoding. 

\subsubsection{Relative Temporal Encoding} \label{subsec:temporal_encoding}

The temporal encoding is to represent the temporal information of each node in the substructure node set. Instead of the absolute encoding in \cite{vaswani2017attention}, we consider a relative encoding for dynamic graphs. Concretely, for each node in the substructure node set $v_j^i \in \mathcal{S}^i(e^t_{\rm tgt})$, the data source for relative time encoding is defined as the difference between the occurring time $t$ of target edge and the current time of timestamp $i$. The motivation behind this is that our main task is to predict the legality of the target edge, so the relative time to the target edge is a more significant factor for anomaly detection. Similar linear mapping is also applied as the encoding function, and the formal expression of relative temporal encoding is given as:

\begin{equation}
\label{eq:temp_enc}
\mathbf{x}_{\rm temp}(v_j^i)= linear(\|t-i\|) \in \mathbb{R}^{d_{enc}},
\end{equation}

\noindent where $\|\cdot\|$ and $linear(\cdot)$ are the relative time computing function and the learnable linear mapping respectively, and $d_{enc}$ is the dimension of node encoding. 

\subsubsection{Encoding Fusion} \label{subsec:encoding_fusion}

After computing the three terms of encoding, we fuse them \HL{as} the input node encoding of the downstream transformer model. For the sake of running efficiency, we define the fused node encoding as the summation of three encoding terms rather than concatenating them into a vector with a higher dimension. The encoding fusion is formalized as follows:

\begin{equation}
\label{eq:fused_enc}
\mathbf{x}(v_j^i)= \mathbf{x}_{\rm diff}(v_j^i) + \mathbf{x}_{\rm dist}(v_j^i) + \mathbf{x}_{\rm temp}(v_j^i) \in \mathbb{R}^{d_{enc}}.
\end{equation}

Finally, given a target edge $e^t_{\rm tgt}$, we calculate the encoding of each node in its substructure node set, and stack them into a encoding matrix which represents the property of $e^t_{\rm tgt}$. The encoding matrix is represented by:

\begin{equation}
\label{eq:concat}
\HL{\mathbf{X}(e^t_{\rm tgt}) = \bigoplus\limits_{v_j^i \in \mathcal{S}(e_{\rm tgt}^t)} [\mathbf{x}(v_j^i)]^\mathsf{T} \in \mathbb{R}^{(\tau (k + 2)) \times d_{enc}},}
\end{equation}

\noindent \HL{where $\bigoplus$ is the concatenation operation and $[\cdot]^\mathsf{T}$ is the transpose operation. }

\subsection{Dynamic Graph Transformer} \label{subsec:transformer}

To learn knowledge from dynamic graphs, the neural network model should consider both of the spatial structure information and temporal dynamic information. In most situations, the spatial information and temporal information are coupled and should be captured simultaneously for efficient anomaly detection. Taking the dynamic graph in Figure \ref{fig:framework} as an example, the node $v^t_1$ and $v^t_2$ have an connection at time $t$, a precursor is that their communities have several connections in the previous timestamps, e.g., $u^{t-1}_1$-$u^{t-1}_2$ and $u^{t-2}_1$-$u^{t-2}_2$. For the design of dynamic graph encoder, a question that arises is: \textit{how can a neural network-based encoder consider the spatial and temporal information simultaneously?} A general solution in the existing works is to use hybrid networks stacked by spatial module and temporal module. The spatial/temporal modules are employed in such hybrid models to capture spatial/temporal information respectively and separately. For instance, in StrGNN, the GCN serves as a spatial module, and GRU processes the output of GCN from different timestamps to capture temporal information. A limitation of such hybrid models is that they may miss some information that crosses spatial and temporal domains, which further leads to a sub-optimal solution.

To learn the spatial and temporal knowledge in the dynamic graphs, we propose to adopt a transformer model solely as the encoder. With the multiple timestamps of node encoding as input, the \textit{dynamic graph transformer} can simultaneously capture both spatial and temporal features with a single encoder. The dynamic graph transformer is composed of two modules: the transformer module and the pooling module. With the transformer module, the abundant cross-domain knowledge is captured by the attention mechanism, and the final attention layer generates the informative node embeddings. After that, the \HL{pooling} module aggregates the embedding of all nodes in the substructure node set into an embedding vector for the target edge.

\subsubsection{Transformer Module}
The target of the transformer module is to aggregate the \HL{encodings of nodes within a substructure node set into node embeddings.} 
\HL{To this end, a number} of attention layers are \HL{utilized to exchange the information of different nodes. To be concrete, a} single \HL{attention} layer can be written as:

\begin{equation}
\label{eq:att_layer}
\begin{aligned}
\mathbf{H}^{(l)} =attention\left(\mathbf{H}^{(l-1)}\right)=softmax\left(\frac{\mathbf{Q}^{(l)} \mathbf{K}^{(l)\top}}{\sqrt{d_{emb}}}\right) \mathbf{V}^{(l)},
\end{aligned}
\end{equation}

\noindent where $\mathbf{H}^{(l)}$ and $\mathbf{H}^{(l-1)}$ is the output embedding of the $l$ and $(l-1)$ -th layer, $d_{emb}$ is the dimension of node embedding, and $\mathbf{Q}^{(l)}$, $\mathbf{K}^{(l)}$, $\mathbf{V}^{(l)} \in \mathbb{R}^{(\tau (k + 2)) \times d_{emb}}$ are the query matrix, key matrix and value matrix \HL{for feature transformation and information exchange}. Concretely, the $\mathbf{Q}^{(l)}$, $\mathbf{K}^{(l)}$ and $\mathbf{V}^{(l)}$ are computed by:

\begin{equation}
\label{eq:att_weight}
\left\{\begin{aligned}
\mathbf{Q}^{(l)} &=\mathbf{H}^{(l-1)} \mathbf{W}_{Q}^{(l)}, \\
\mathbf{K}^{(l)} &=\mathbf{H}^{(l-1)} \mathbf{W}_{K}^{(l)}, \\
\mathbf{V}^{(l)} &=\mathbf{H}^{(l-1)} \mathbf{W}_{V}^{(l)},
\end{aligned}\right.
\end{equation}

\noindent where $\mathbf{W}_{Q}^{(l)}, \mathbf{W}_{K}^{(l)}, \mathbf{W}_{V}^{(l)} \in \mathbb{R}^{d_{emb} \times d_{emb}}$ are the learnable parameter matrices of the $l$-th attention layer. \HL{In an attention layer, $\mathbf{Q}^{(l)}$ and $\mathbf{K}^{(l)}$ calculate the contributions of different nodes' embeddings, while $\mathbf{V}^{(l)}$ projects the input into a new feature space. Equation (\ref{eq:att_layer}) combines them and acquires the output embedding of each node by aggregating the information of all nodes adaptively.}

\HL{In our transformer module}, the input of the transformer module $\mathbf{H}^{(0)}$ is defined as the encoding matrix of the target edge $\mathbf{X}(e^t_{\rm tgt})$, and here we simply set $d = d_{emb} = d_{enc}$ to align the dimension. The output of the final attention layer $\mathbf{H}^{(L)}$ is denoted as the output node embedding matrix $\mathbf{Z}$ of the transformer module, where each row represents an embedding vector of the corresponding node.

\subsubsection{\HL{Pooling} Module}

The target of pooling module is to transfer the embeddings of nodes in substructure $\mathbf{Z}$ into a target edge embedding vector $\mathbf{z}(e_{\rm tgt}^t)$. Here we utilize \HL{the} average pooling \HL{operation} as \HL{our pooling} function, which has been applied in previous works \cite{liu2021anomaly}. The \HL{pooling} function is formalized as:

\begin{equation}
\label{eq:pooling}
\mathbf{z}(e_{\rm tgt}^t) = pooling(\mathbf{Z})  = \sum_{k=1}^{n_s} \frac{\left(\mathbf{Z}\right)_{k}}{n_s},
\end{equation}

\noindent where $\left(\mathbf{Z}\right)_{k}$ is the $k$-th row of $\mathbf{Z}$, and $n_s=\tau(k+2)$ is the number of nodes of the substructure node set $\mathcal{S}(e_{\rm tgt}^t)$. 

\subsection{Discriminative Anomaly Detector} \label{subsec:detector}

After the edge embedding is acquired, the target of anomaly detection is to learn an anomaly score for each edge in the dynamic graph. Here, we consider an end-to-end framework where a neural network-based anomaly detector computes the anomaly score. However, in our learning setting, there is not any ground-truth anomaly sample in the training set. Such \HL{a} situation brings a new challenge: \textit{How to learn an anomaly detector without any given anomalous sample?} Our solution is to generate pseudo anomalous edges via a negative sampling strategy and train the anomaly detector with the existing edges in the training set (positive edges) as well as the pseudo anomalous edges (negative edges) together.

A simple negative sampling strategy is performed in our framework. For each timestamp of graph whose number of edges is $m^t$, we randomly sample the equal number of node pairs as the candidates of negative pairs. Then, we check all these node pairs to ensure that they do not belong to the existing normal edge set in all the training timestamps. We resample a new pair and perform validation for each illegal node pair until the node pair is valid. After negative sampling, we use context sampling to acquire the substructure node set of each negative edge and perform spatial-temporal node encoding. Then, the encoding is fed into the dynamic graph transformer model to obtain the embedding of the negative edge. 

The anomaly detector is constructed to discriminate the positive and negative edge embeddings. A fully connected neural network layer with Sigmoid activation is served as the scoring module which computes the anomaly scores of edge embeddings, which is formalized by

\begin{equation}
\label{eq:scoring}
f(e) = Sigmoid\big(\mathbf{z}(e) \mathbf{w}_S + b_S\big)
\end{equation}

\noindent where $f(e)$ and $\mathbf{z}(e)$ is the anomaly score and edge embedding of edge $e$ respectively, $Sigmoid(\cdot)$ is the Sigmoid activation function, $\mathbf{w}_S \in \mathbb{R}^{d_{emb}} $ and $b_S \in \mathbb{R}$ are the weights and bias parameters of fully connected neural network layer respectively.

\begin{algorithm}[t]
	\caption{The Overall Training Procedure of TADDY}
	\label{algo:framework}
	\renewcommand{\algorithmicrequire}{\textbf{Input:}}
	\renewcommand{\algorithmicensure}{\textbf{Output:}}
	\begin{algorithmic}[1]
		\REQUIRE {Training set  of dynamic graph: $\mathbb{G}=\{\mathcal{G}^t\}^{T}_{t=1}$, Number of training epochs: $I$, Number of sampled contextual nodes: $k$, Size of time window: $\tau$}.
		\STATE Randomly initialize the parameters of encoding linear mappings, transformer model and scoring function
		\FOR{$i \in 1,2,\cdots,I$}
		\FOR{timestamp $\mathcal{G}^t=(\mathcal{V}^t,\mathcal{E}^t) \in \{\mathcal{G}^t\}^{T}_{t=\tau}$}
		\STATE Sample negative edge set $\mathcal{E}^t_n \sim P_{n}(\mathcal{E}^t)$ by negative sampling strategy
		\FOR{$e \in \mathcal{E}^t \cup \mathcal{E}^t_n$}
		
		\STATE Set $e$ as the target edge and sample its substructure node set $\mathcal{S}(e)$ with $\tau(k+2)$ nodes
		\STATE Calculate node encoding matrix $\mathbf{X}(e)$ via Equation (\ref{eq:diff_enc}) - (\ref{eq:fused_enc})
		\STATE Calculate edge embedding vector $\mathbf{z}(e)$ via Equation (\ref{eq:att_layer}) - (\ref{eq:pooling})
		\STATE Calculate anomaly score $f(e)$ via Equation (\ref{eq:scoring})
		\ENDFOR
		\STATE Calculate loss function $\mathcal{L}$ via Equation (\ref{eq:bce})
		\STATE Back propagation and update the parameters
		\ENDFOR
		\ENDFOR
		
	\end{algorithmic}
\end{algorithm}

We employ a binary cross-entropy loss function with pseudo labels to train the framework in an end-to-end manner. For the positive edges, we expect them to have a small anomaly score, hence their pseudo label is $0$; for the negative edges, conversely, the pseudo label is $1$. For a training timestamp $\mathcal{G}^t=(\mathcal{V}^t,\mathcal{E}^t)$ whose edge number is $m^t$, the loss function is given as

\begin{equation}
\label{eq:bce}
\mathcal{L} = -\sum_{i=1}^{m^t} \log \big( 1 - f(e_{{\rm pos},i})\big)+ \log \big(f(e_{{\rm neg},,i})\big)
\end{equation}

\noindent where $e_{{\rm pos},i} \in \mathcal{E}^t$ is the $i$-th positive edge and $e_{{\rm neg},,i} \in \mathcal{E}^t_n \sim P_{n}(\mathcal{E}^t)$ is the $i$-th negative edge sampled by the negative sampling strategy.

To the end, the overall training procedure of our TADDY framework is depicted in Algorithm \ref{algo:framework}. The framework is trained in an iterative and end-to-end manner. In each iteration, different negative edges are sampled to prevent training bias and over-fitting. For all positive and negative edges, we perform substructure sampling, node encoding, transformer embedding and anomaly score computing sequentially. Finally, the parameters are updated by back propagation under the supervision of binary cross-entropy loss function. When the framework is well trained, the anomaly scores for test edges can be obtained by executing the line 6-9 in Algorithm \ref{algo:framework}.

\subsection{Complexity Analysis} \label{subsec:complexity}
In this subsection, we analyze the time complexity of each component in TADDY framework. For edge-based substructure sampling, the complexity is mainly caused by the computation of graph diffusion, which is $\mathcal{O}(T {\widetilde{n}}^2)$ where $\widetilde{n}$ is the average number of nodes for graph timestamps and $T$ is the number of timestamps. In spatial-temporal node encoding and dynamic graph transformer, we process all nodes in the substructure node set for each target edge, which brings a complexity of $\mathcal{O}(\tau k)$ for one edge. Therefore, for $I$ iterations, the total time complexity is $\mathcal{O}(\tau k m I)$ where $m$ is the number of edges. For discriminative anomaly detector, the time complexity is $\mathcal{O}(m I)$, which is far less than the other components and can be ignored. To sum up, the overall time complexity is $\mathcal{O}(\tau k m I + T {\widetilde{n}}^2)$.

\section{Experiments} \label{sec:exp}

In this section, we evaluate the performance of the proposed TADDY via extensive experimental \HL{studies}. We first introduce the setup for our experiments. We demonstrate the experimental results in the rest three subsections, including performance comparison, parameter study, and ablation study.

\subsection{Experimental Setup}

\subsubsection{Datasets}

\begin{table}
	\centering
	\caption{The statistics of the datasets. \HL{For each dataset, the number of nodes, the number of edges, and the average degree are reported.}}
	\vspace{-2mm}
	\begin{tabular}{@{}c|c|c|c@{}}
		\toprule
		{Dataset} &  {$\sharp$ nodes} & {$\sharp$ edges} & {Avg. Degree} \\ \midrule
		{UCI Messages}  & 1,899    & 13,838      & 14.57   \\
		{Digg}          & 30,360   & 85,155      & 5.61    \\
		\HL{Email-DNC}     & \HL{1,866}    & \HL{39,264}      & \HL{42.08}   \\
		{Bitcoin-Alpha} & 3,777    & 24,173      & 12.80   \\
		{Bitcoin-OTC}   & 5,881    & 35,588      & 12.10   \\
		\HL{AS-Topology}   & \HL{34,761}   & \HL{171,420}     & \HL{9.86}    \\
		
		\bottomrule
	\end{tabular}
	\label{table:dataset}
	\vspace{-2mm}
\end{table}

We evaluate our proposed TADDY framework on \HL{six} 
real-world benchmark datasets of dynamic graphs. The statistics of the datasets are given in Table \ref{table:dataset}, and the detailed descriptions are demonstrated as follows.

\textbf{UCI Messages}\footnote{http://konect.cc/networks/opsahl-ucsocial} \cite{opsahl2009clustering} is a social network dataset collected from an online community of students at University of California, Irvine. In the constructed dynamic graph, each node indicates a user, and each edge represents a message between two users.  

\textbf{Digg}\footnote{http://konect.cc/networks/munmun\_digg\_reply} \cite{de2009social} is a network dataset collected from a news website digg.com. In Digg dataset, each node is a website user, and each edge indicates that one user replies to another user. 

\HL{\textbf{Email-DNC}\footnote{http://networkrepository.com/email-dnc} \cite{nr} is network of emails in the 2016 Democratic National Committee email leak. Each node corresponds to a person in the United States Democratic Party, and each edge denotes that a person has sent an email to another person.}

\textbf{Bitcoin-Alpha}\footnote{http://snap.stanford.edu/data/soc-sign-bitcoin-alpha} and \textbf{Bitcoin-OTC}\footnote{http://snap.stanford.edu/data/soc-sign-bitcoin-otc} \cite{kumar2016edge,kumar2018rev2} are two who-trusts-whom networks of bitcoin users trading on the platforms from www.btc-alpha.com and www.bitcoin-otc.com respectively. In these two datasets, the nodes are the users from the platform, and an edge appears when one user rates another on the platform.

\HL{\textbf{AS-Topology}\footnote{http://networkrepository.com/tech-as-topology} \cite{zhang2005collecting} is a network connection dataset collected from autonomous systems of the Internet. Each node in the graph corresponds to an autonomous system, and each edge indicates a connection between two autonomous systems.}

We pre-process the datasets following previous works \cite{yu2018netwalk,zheng2019addgraph}. The edges in each dataset are annotated with timestamps. The repeated edges in the edge stream are removed in the pre-processing phase. Since there is no ground-truth anomalous edge in the original datasets, we follow the approach used in \cite{yu2018netwalk} to inject anomalous edges in each dataset. \HL{To be concrete, the training data is totally clean. For each snapshot $\mathcal{G}^t$ in the test set, we randomly link $p_A \times m^t$ pairs of disconnected nodes as anomalous edges, where $p_A$ is the anomaly proportion indicating the percentage of anomalous edges in each snapshot, and $m^t$ is the (original) number of edges in $\mathcal{G}^t$.}

\begin{table*}[t]

  \centering
  \caption{Anomaly detection performance comparison \HL{reported in AUC measure. The upper three baselines belong to graph embedding methods, and the middle three baselines belong to deep dynamic graph anomaly detection methods.} The best performing method in each experiment is in bold.}
  \vspace{-2mm}
  \label{table:results}
  \begin{tabular}{l|p{30 pt}<{\centering}p{30 pt}<{\centering}p{30 pt}<{\centering}|p{30 pt}<{\centering}p{30 pt}<{\centering}p{30 pt}<{\centering}|p{30 pt}<{\centering}p{30 pt}<{\centering}p{30 pt}<{\centering}}
    \bottomrule
    \multirow{2}*{Methods} &
    \multicolumn{3}{c|}{UCI Messages} & \multicolumn{3}{c|}{Digg} & \multicolumn{3}{c}{\HL{Email-DNC}}\\
    \cline{2-10}
    & 1\% & 5\% & 10\% & 1\% & 5\% & 10\% & 1\% & 5\% & 10\%\\
    \hline
node2vec            & 0.7371 & 0.7433 & 0.6960 & 0.7364 & 0.7081 & 0.6508 & 0.7391 & 0.7284 & 0.7103\\
Spectral Clustering & 0.6324 & 0.6104 & 0.5794 & 0.5949 & 0.5823 & 0.5591 & 0.8096 & 0.7857 & 0.7759\\
DeepWalk            & 0.7514 & 0.7391 & 0.6979 & 0.7080 & 0.6881 & 0.6396 & 0.7481 & 0.7303 & 0.7197\\
    \hline
NetWalk             & 0.7758 & 0.7647 & 0.7226 & 0.7563 & 0.7176 & 0.6837 & 0.8105 & 0.8371 & 0.8305\\
\HL{AddGraph}            & 0.8083 & 0.8090 & 0.7688 & 0.8341 & 0.8470 & 0.8369 & 0.8393 & 0.8627 & 0.8773\\
StrGNN              & 0.8179 & 0.8252 & 0.7959 & 0.8162 & 0.8254 & 0.8272 & 0.8775 & 0.9103 & 0.9080\\
    \hline
TADDY               & \bf{0.8912} & \bf{0.8398} & \bf{0.8370} & \bf{0.8617} & \bf{0.8545} & \bf{0.8440} & \bf{0.9348} & \bf{0.9257} & \bf{0.9210} \\
\toprule
\bottomrule
      \multirow{2}*{Methods} &
    \multicolumn{3}{c|}{Bitcoin-Alpha} & \multicolumn{3}{c|}{Bitcoin-OTC} & \multicolumn{3}{c}{\HL{AS-Topology}} \\
    \cline{2-10}
    & 1\% & 5\% & 10\% & 1\% & 5\% & 10\% & 1\% & 5\% & 10\%\\
    \hline
node2vec            & 0.6910 & 0.6802 & 0.6785 & 0.6951 & 0.6883 & 0.6745 & 0.6821 & 0.6752 & 0.6668\\
Spectral Clustering & 0.7401 & 0.7275 & 0.7167 & 0.7624 & 0.7376 & 0.7047 & 0.6685 & 0.6563 & 0.6498\\
DeepWalk            & 0.6985 & 0.6874 & 0.6793 & 0.7423 & 0.7356 & 0.7287 & 0.6844 & 0.6793 & 0.6682\\
    \hline
NetWalk             & 0.8385 & 0.8357 & 0.8350 & 0.7785 & 0.7694 & 0.7534 & 0.8018 & 0.8066 & 0.8058 \\
\HL{AddGraph}            & 0.8665 & 0.8403 & 0.8498 & 0.8352 & 0.8455 & 0.8592 & 0.8080 & 0.8004 & 0.7926 \\
StrGNN              & 0.8574 & 0.8667 & 0.8627 & 0.9012 & 0.8775 & 0.8836 & 0.8553 & 0.8352 & 0.8271 \\
    \hline
TADDY               & \bf{0.9451} & \bf{0.9341} & \bf{0.9423} & \bf{0.9455} & \bf{0.9340} & \bf{0.9425} & \bf{0.8953} & \bf{0.8952} & \bf{0.8934} \\
    \toprule
  \end{tabular}
  \vspace{-2mm}
\end{table*}

\subsubsection{Baselines}

We compared TADDY framework against \HL{six} state-of-the-art baselines that can be categorized into two groups: graph embedding methods and deep dynamic graph anomaly detection methods.

\textbf{DeepWalk} \cite{perozzi2014deepwalk} is a random walk-based method for graph embedding. It generates random walks with a given length starting from a target node and uses a Skip-gram-like manner to learn embedding for unattributed graphs.

\textbf{node2vec} \cite{grover2016node2vec} considers breadth-first traversal and depth-first traversal when generating random walks. The Skip-gram technology is also employed to learn node embedding in node2vec.

\textbf{Spectral Clustering} \cite{von2007tutorial} learns node embedding by maximizing the similarity between bodes in neighborhood. The intuition behind this method is to preserve the local connection relationship in graphs.

\textbf{NetWalk} \cite{yu2018netwalk} is a representative anomaly detection method for dynamic graph. It utilizes a random walk-based approach to generate contextual information and learns node embedding with an auto-encoder model. The node embeddings are updated incrementally over time via a reservoir-based algorithm. The anomaly is detected using a dynamic-updated clustering on the learned embedding.

\HL{\textbf{AddGraph} \cite{zheng2019addgraph} is an end-to-end dynamic graph anomaly detection approach. It leverages a GCN module to capture spatial information, and employs a GRU-attention module to extract short- and long- term dynamic evolving.}

\textbf{StrGNN} \cite{cai2020structural} is an end-to-end graph neural network model for detecting anomalous edges in dynamic graphs. It leverages an h-hop enclosing subgraph as the network's input and combines GCN and GRU to learn structural-temporal information for each edge. 

For the graph embedding methods, the K-means clustering-based anomaly detector, which is presented in NetWalk \cite{yu2018netwalk} is utilized to detect anomalies based on the learned node embeddings.

\subsubsection{Experimental Design} 
In our experiments, each dataset is divided into two subsets: the first $50\%$ of timestamps is denoted as training set, while the latter $50\%$ as test set. We consider three different anomaly proportions \HL{$p_A$}, $1\%$, $5\%$, and $10\%$, when injecting the anomalous data into the test set. To measure the performance of the proposed framework as well as the baselines, ROC-AUC (AUC for short) is employed as our primary metric. The ROC curve indicates a plot of true positive rate against false positive rate where anomalous labels are viewed as ``positive''. AUC is defined as the area under the ROC curve, which indicates the probability that a randomly selected anomalous edge is ranked higher than a normal edge. The value range of AUC is $0$ to $1$, and a larger value represents a \HL{better} anomaly detection performance. 

\subsubsection{Parameter Settings} 
All the parameters can be tuned by 5-fold cross-validation on a rolling basis. For edge-based substructure sampling, we set the number $k$ of contextual nodes to be $5$ and $\tau$ is selected from $1$ to $4$. We use PPR diffusion in \HL{our experiments}, which is computed by Eq. (\ref{eq:ppr}). For spatial-temporal node encoding, the dimension of encoding $d_{enc}$ is $32$, which is the same as $d_{emb}$ in Dynamic Graph Transformer. The number of attention layers is $2$ for all the datasets, and the number of attention heads is $2$. The framework is trained by Adam optimizer with a learning rate of $0.001$. We train UCI Messages, Bitcoin-Alpha, and Bitcoin-OTC datasets with $100$ epochs and \HL{the rest three} datasets for $200$ epochs. The snapshot size is set to be $1,000$ for UCI Messages and Bitcoin-OTC, $2,000$ for \HL{Email-DNC and} Bitcoin-Alpha, and $6,000$ for Digg \HL{and AS-Topology}, respectively. 

\subsubsection{\HL{Computing Infrastructures}} 
\HL{The proposed method is implemented using PyTorch 1.7.1 \cite{paszke2019pytorch}. 
All experiments are conducted on a computer server with four Quadro RTX 6000 (24GB memory each) GPUs, an Intel Xeon Silver 4214R (2.40 GHz) CPU and 64 GB of RAM.}

\subsection{Anomaly Detection Results}

In this subsection, we report anomaly detection performance and compare our proposed TADDY framework with the baseline methods. The anomaly detection performance comparison of average AUC on all test timestamps is demonstrated in Table \ref{table:results}. We summarize the following observations for the results:

\begin{itemize} 
	\item The proposed TADDY framework consistently outperforms all baselines on the \HL{six} 
	dynamic graph datasets with different anomaly proportions. Compared to the baseline method with the best results, TADDY reaches a performance gain of \HL{$4.49\%$} on AUC averagely. The main reason is that TADDY extracts the spatial-temporal information by constructing informative node encoding and captures structural dynamic and temporal dynamic simultaneously with a transformer encoder. 
	
	\item Compared to \HL{three} graph embedding-based methods, the deep dynamic graph anomaly detection methods, NetWalk, \HL{AddGraph}, StrGNN and TADDY, always have a more competitive performance. We attribute this performance advantage to the leverage of temporal information. By considering the interaction in previous timestamps, these methods learn the dynamic evolving in graphs.
	
	\item TADDY has a larger performance gain when the \HL{anomalies} are scarce. Concretely, the \HL{average} performance gap \HL{on AUC} between TADDY and the best baseline under $1\%$ anomaly proportion is \HL{$5.35\%$}, while \HL{the gaps under $5\%$ and $10\%$ anomaly proportions are $3.69\%$ and $4.43\%$, respectively.} A possible reason is that we train the framework with an efficient negative sampling strategy, which ensures the robustness under different anomaly proportion in the test set. 
	
	\item On the two Bitcoin datasets, TADDY achieves more remarkable results. \HL{Compared to the best baseline,} the average \HL{AUC} gain on Bitcoin-Alpha and Bitcoin-OTC is \HL{$6.42\%$}, which is \HL{significantly} higher than the \HL{AUC} gain on the rest datasets (\HL{$3.53\%$}). The reason for such remarkable advantages is that the abnormality of Bitcoin transaction is more closely related to spatial-temporal dynamic, and TADDY can successfully capture such dynamic by comprehensive node embedding and attention mechanism. 
	
\end{itemize}

\begin{figure*}[ht]
	\centering
	\vspace{-3mm}
		\subfigure[UCI Messages]{
		\includegraphics[width=5cm]{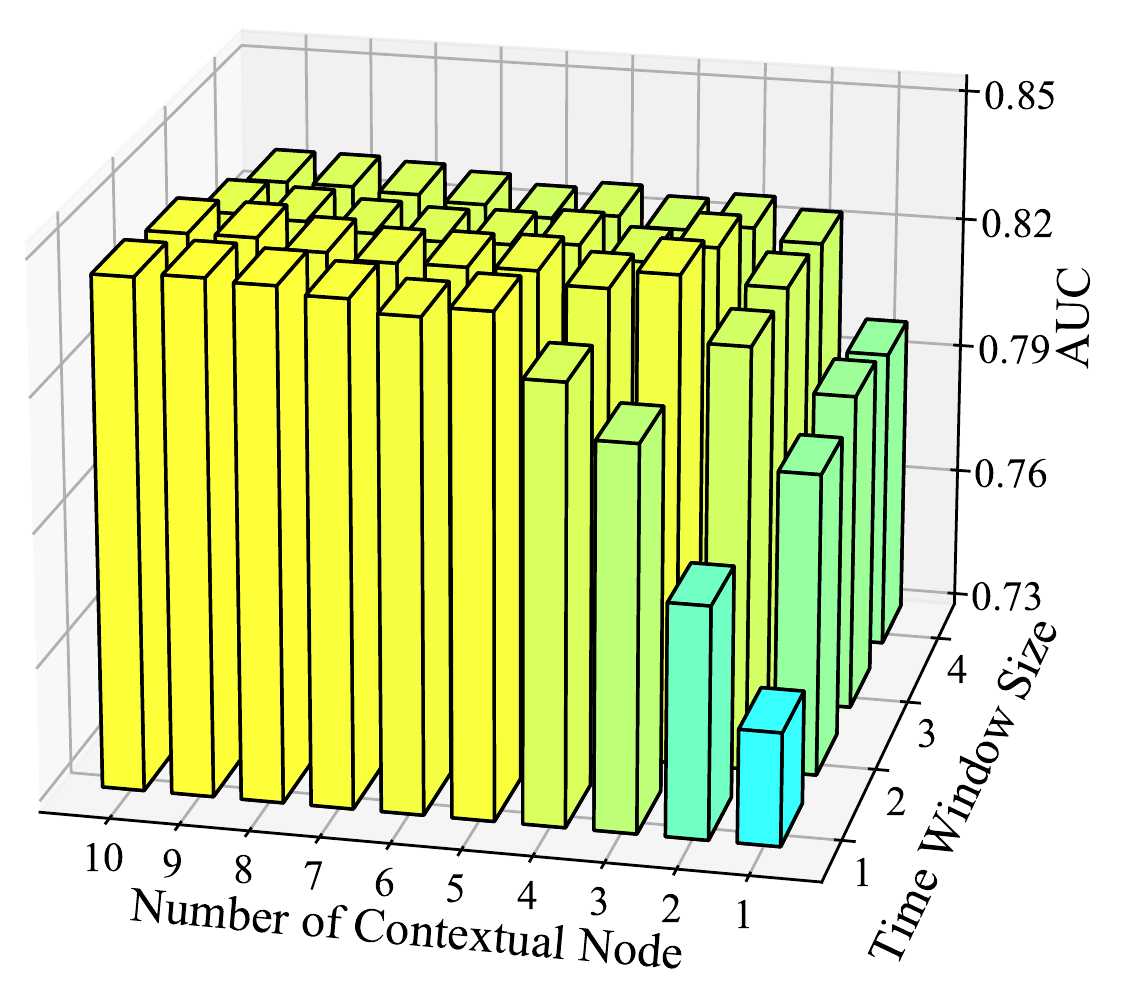}
	}\hfill
			\subfigure[Bitcoin-Alpha]{
		\includegraphics[width=5cm]{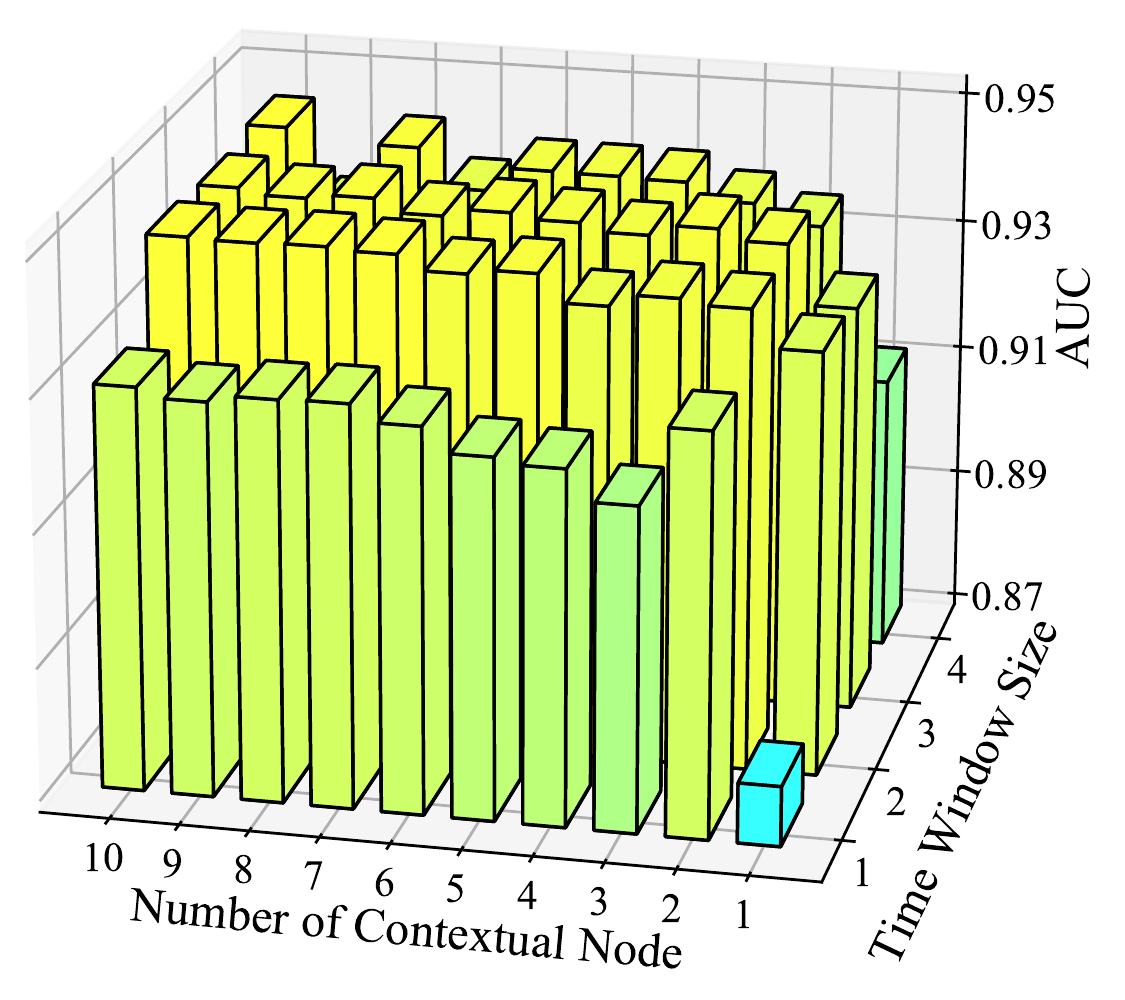}
	}\hfill
			\subfigure[Bitcoin-OTC]{
		\includegraphics[width=5cm]{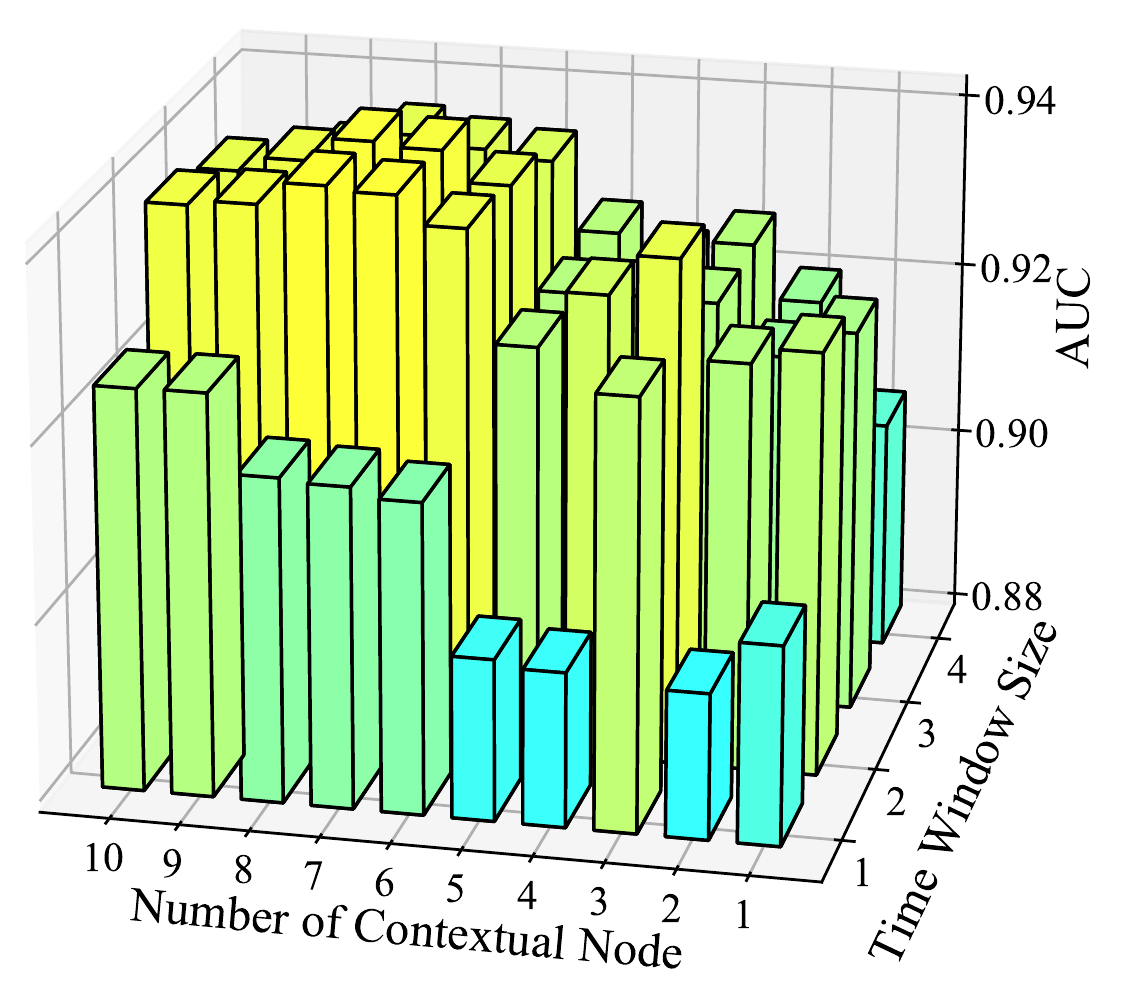}
	}
	\vspace{-3mm}
	\caption{The sensitivity of contextual node number $k$ and time window size $\tau$ on three datasets. \HL{The vertical axis represents the AUC values of TADDY with different $k$ and $\tau$.} A warmer color indicates a higher AUC value.}
	\label{fig:param_kt}
\end{figure*}

\begin{figure*}[ht]
	\centering
	\vspace{-3mm}
		\subfigure[UCI Messages]{
		\includegraphics[width=5cm]{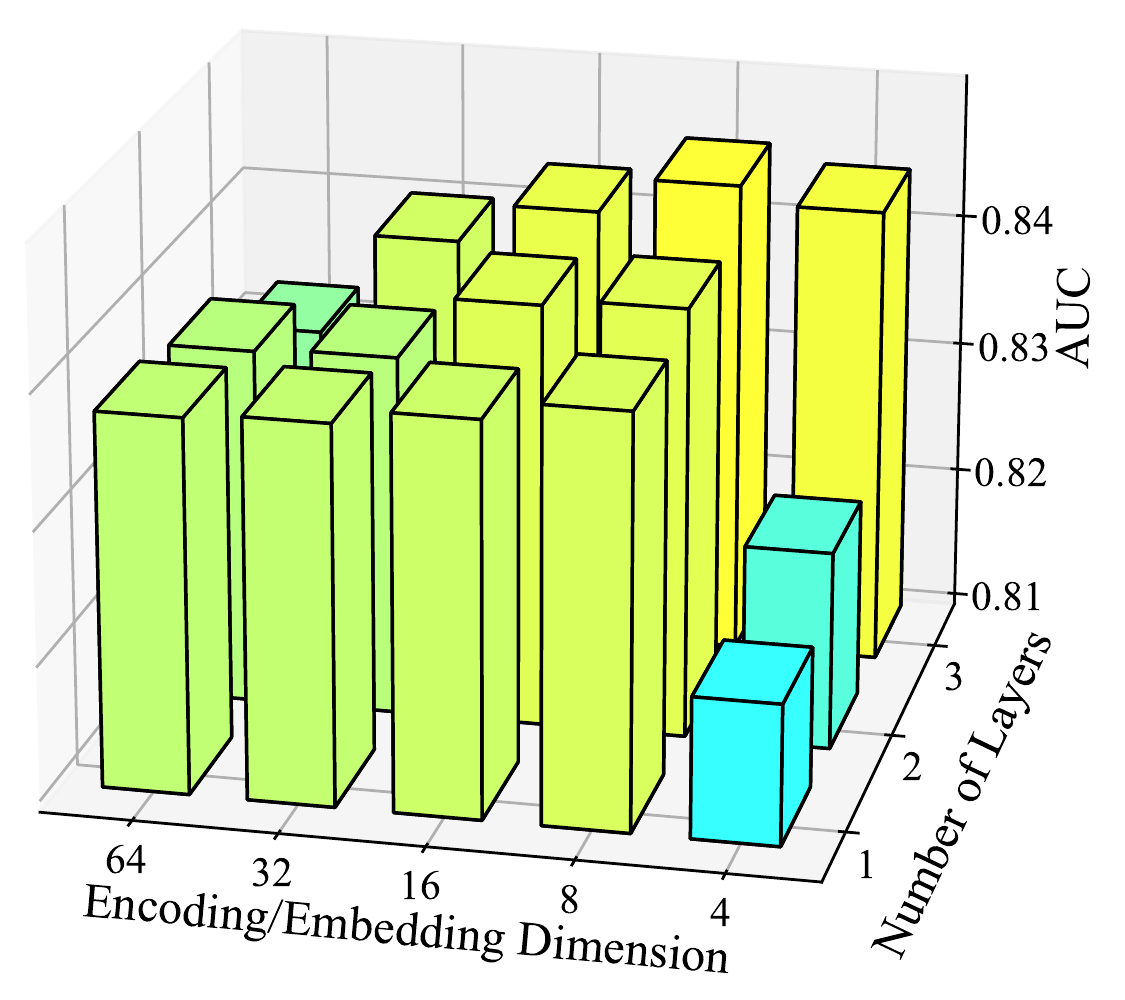}
	}\hfill
			\subfigure[Bitcoin-Alpha]{
		\includegraphics[width=5cm]{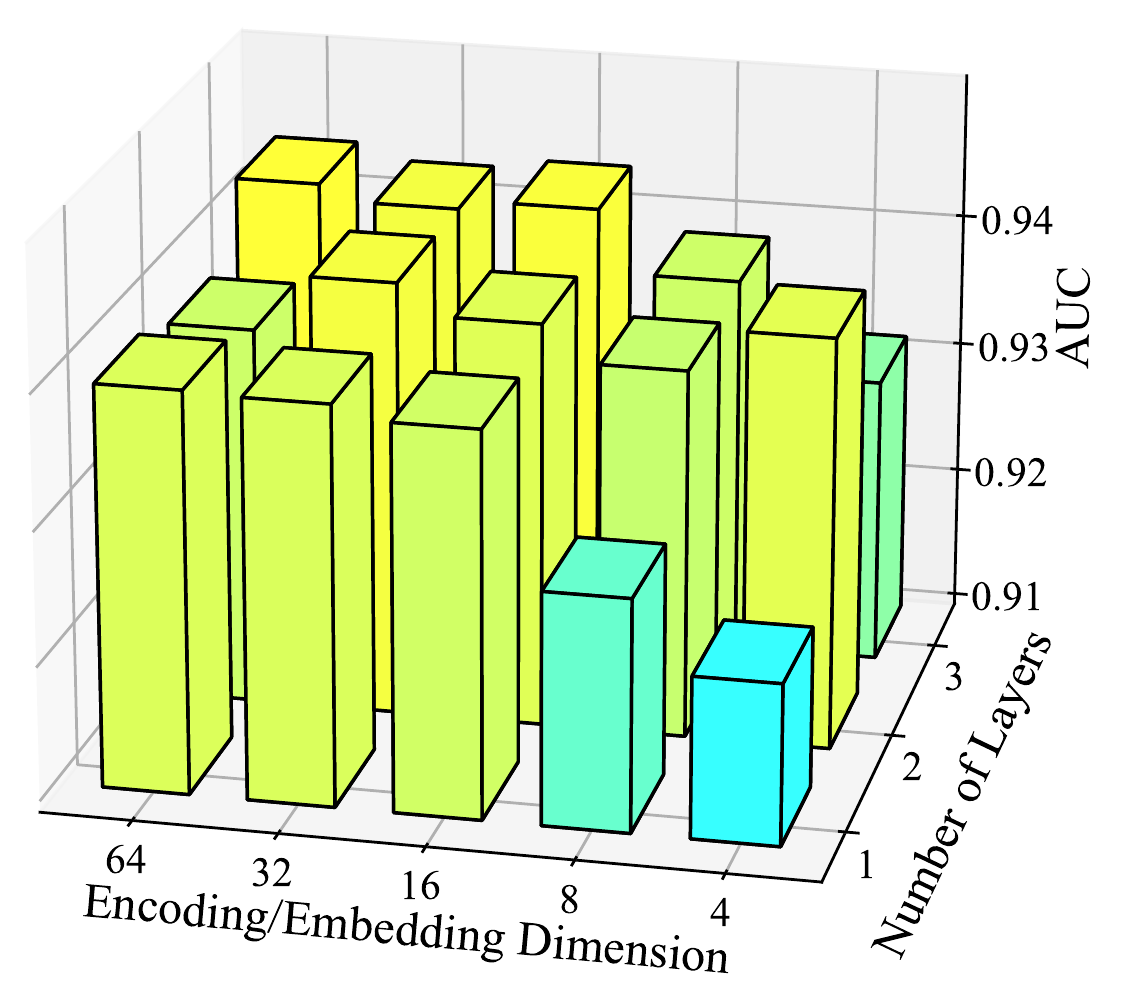}
	}\hfill
			\subfigure[Bitcoin-OTC]{
		\includegraphics[width=5cm]{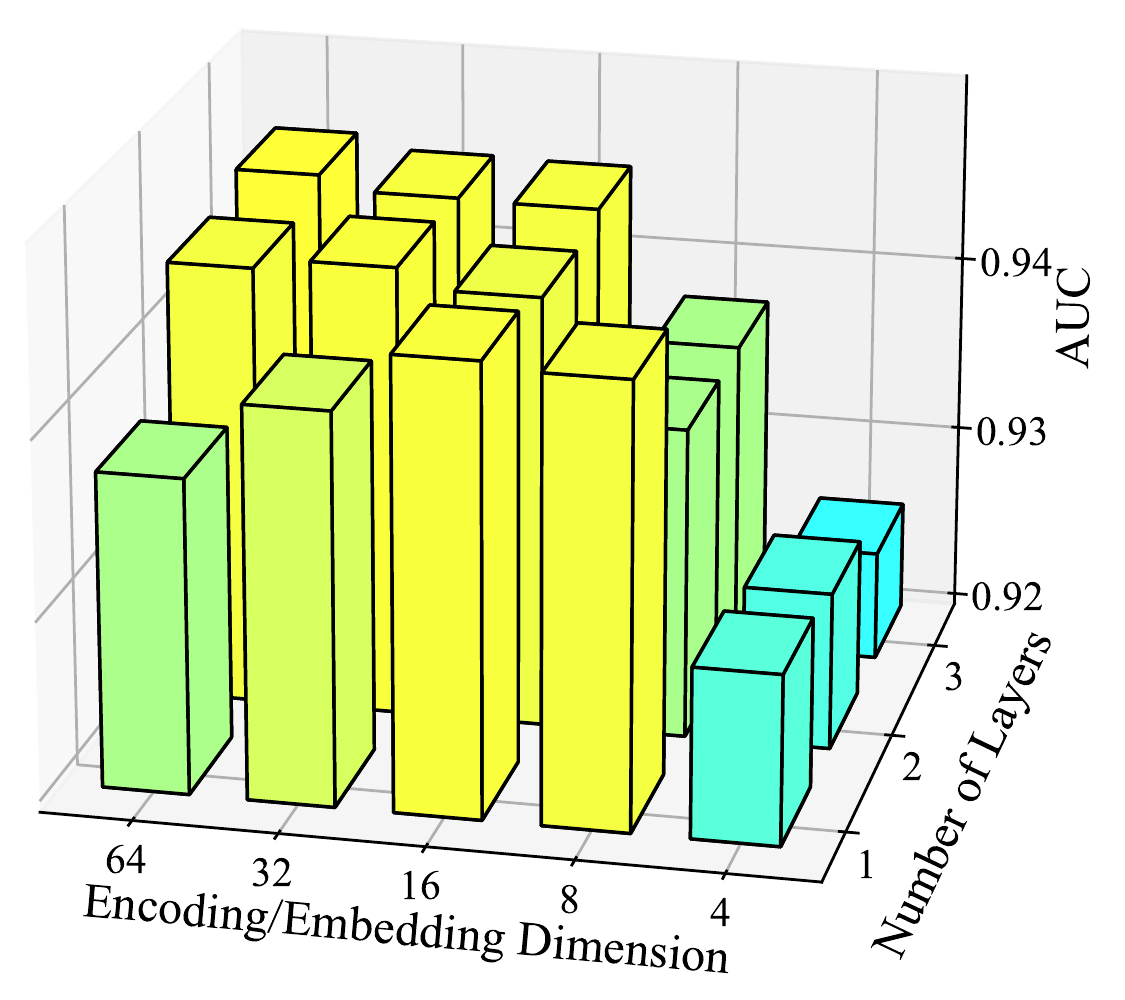}
	}
	\vspace{-3mm}
	\caption{The sensitivity of encoding/embedding dimension $d$ and the number of layers $L$ on three datasets. \HL{The vertical axis represents the AUC values of TADDY with different $d$ and $L$.} A warmer color indicates a higher AUC value.}
	\label{fig:param_dl}
\end{figure*}

\subsection{Parameter Sensitivity}

In this subsection, we investigate the influence of hyper-parameters on TADDY, including the number of contextual nodes and time window size in edge-based substructure sampling, the dimension of encoding/embedding and the number of attention layers in dynamic graph transformer, and the ratio of training data. Here we carry out the experiments on three datasets \HL{(UCI Messages, Bitcoin-Alpha and Bitcoin-OTC)}. In these experiments, we keep the other parameter as default, and the performance is examined under a $10\%$ anomaly proportion setting.

\subsubsection{Parameters of Edge-based Substructure Sampling} \label{subsubsec: param_kt}

To evaluate the effect of number of contextual nodes $k$ and time window size $\tau$ in edge-based substructure sampling, we set the range of $k$ to $\{1, 2, 3, 4, 5, 6, 7, 8, 9, 10\}$ and the range of $\tau$ to $\{1,2,3,4\}$. The sensitivity of $k$ and $\tau$ is exhibited in Figure \ref{fig:param_kt}. According to the results, we \HL{have} the following observations.

When the contextual node number $k$ is extremely small, the anomaly detection performance is relatively pool. With the growth of $k$, there is a significant boost in AUC. When $k>5$, the detection performance tends to be stable, with $k$ getting larger. The performance trend indicates that contextual node sets with a sufficient size are significant to anomaly detection since the anomalous property of edges highly relies on their neighboring local structure. However, when considering an excessive number of contextual nodes, the performance gain is minor. However, a large $k$ is harmful to the running efficiency due to the linear relationship between time complexity and $k$. Consequently, we fix the value of $k$ to $5$ for the consideration of the trade-off between performance and efficiency.

For different datasets, the appropriate size $\tau$ of the time window is different. For instance, a smaller time window is beneficial to UCI Messages, while the two Bitcoin datasets need a more extended time horizon. The main reason is that the temporal reliance of edges in dynamic graphs highly depends on the datasets. When the \HL{edge} appearance has a long-term dependency on the previous graph evolving, a larger time window is needed to capture the dependency. For other datasets like UCI Messages where anomalies are related to the latest snapshots, a wide time window may result in noisy and redundant input for TADDY framework. Therefore, we select the best $\tau$ value for each dataset in our experiments.

\begin{figure*}[ht]
	\centering
	\vspace{-3mm}
		\subfigure[UCI Messages]{
		\includegraphics[width=5cm]{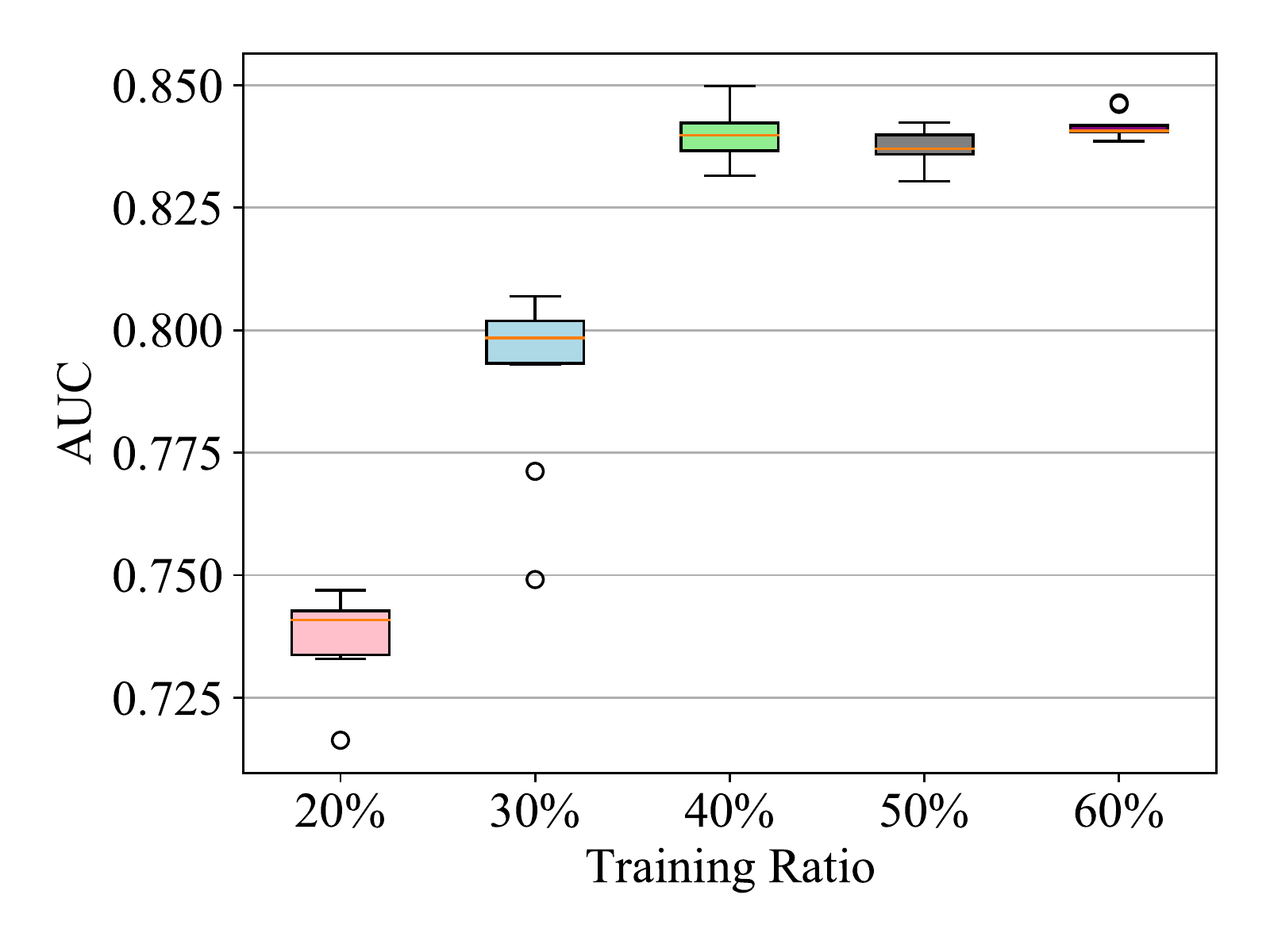}
	}\hfill
			\subfigure[Bitcoin-Alpha]{
		\includegraphics[width=5cm]{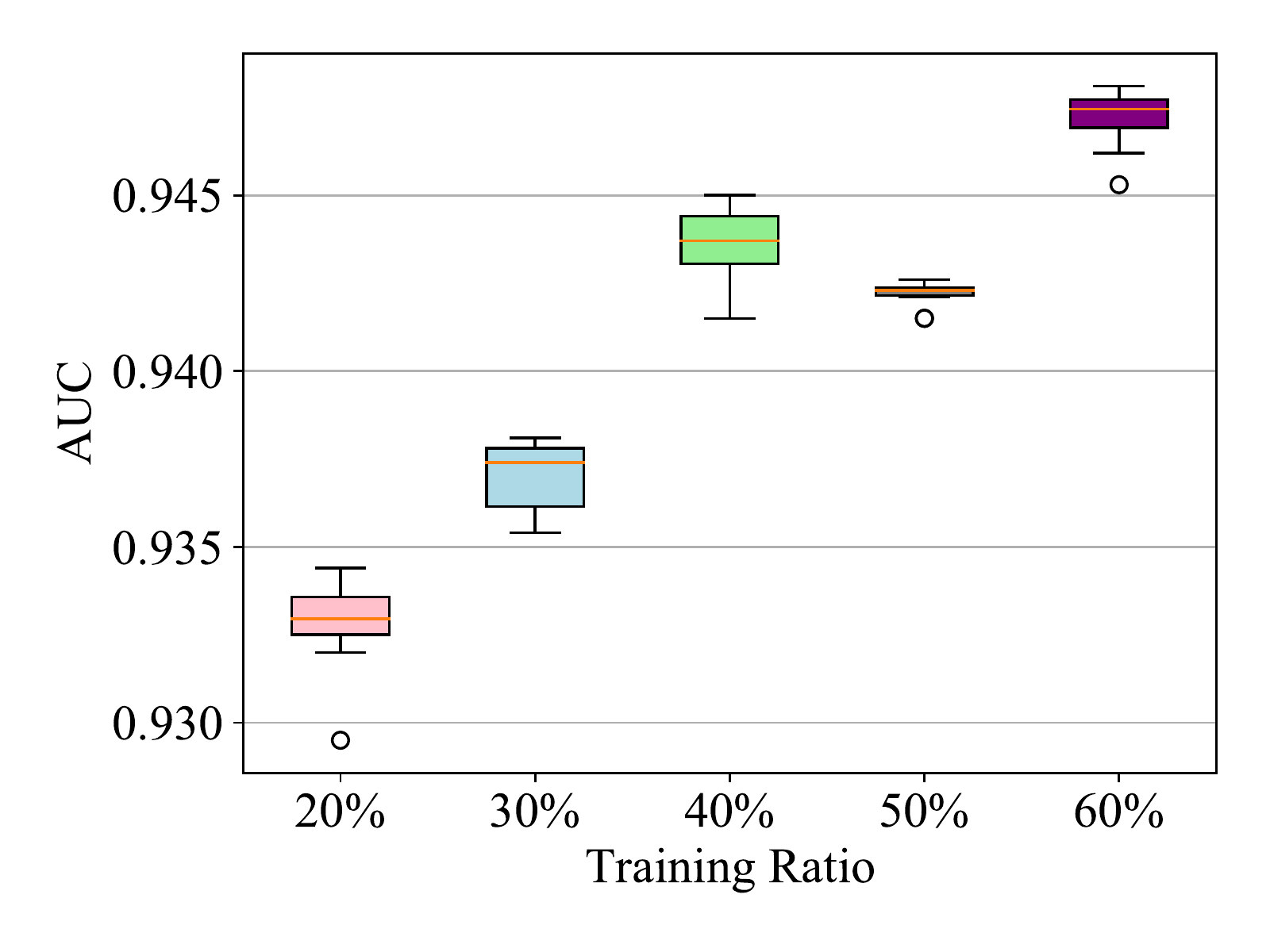}
	}\hfill
			\subfigure[Bitcoin-OTC]{
		\includegraphics[width=5cm]{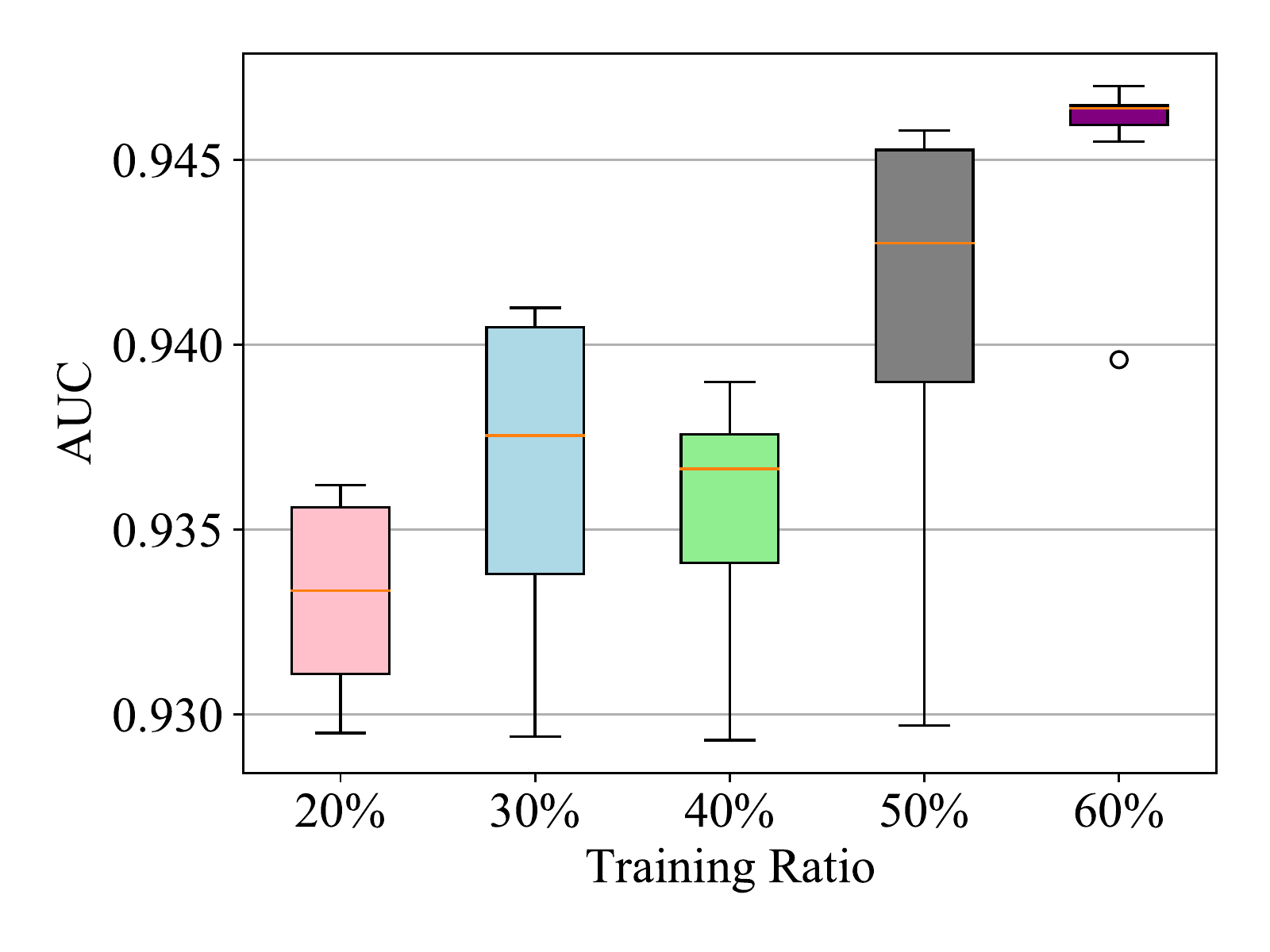}
	}
	\vspace{-3mm}
	\caption{AUC \HL{values of TADDY} on three datasets with different training ratios. The circular markers indicate the results which is viewed as outliers.}
	\label{fig:param_trainrate}
\end{figure*}

\subsubsection{Parameters of Dynamic Graph Transformer} \label{subsubsec: param_dl}

We further investigate the effort of encoding/embedding dimension $d$ and the number of layers $L$ in Transformer model. The value of $d$ is selected from $\{4,8,16,32,64\}$ and which of $L$ is selected from $\{1,2,3\}$. The results are summarized in Figure \ref{fig:param_dl}. 

In Bitcoin-Alpha and Bitcoin-OTC, the AUC \HL{values} increase gradually from $d=4$ to $d=16$, and then go steady after $d=32$. Such observation demonstrates that when $d$ is small, the model may miss useful information. For UCI Messages, $d=8$ seems to be the best choice. When $d$ \HL{getting} larger, there is no significant performance degradation. Our explanation is that when $d$ is too large, the noisy information would be captured by the transformer model.

Compared to $d$, the number of layers has a limited impact on performance. An exception is Bitcoin-Alpha, whose AUC drops when $L=1$, which indicates that a sufficient number of layers can bring adequate interaction among structuring nodes. As such, we fix $L=2$ for each dataset to balance the running speed and detection performance. 

\subsubsection{Training ratios} \label{subsubsec: param_trainrate}

In this experiment, we discuss the performance of TADDY framework using training data with different ratios. The range of training ratio is $\{20\%, 30\%, 40\%, 50\%, 60\%\}$ and other parameters are set to default. Figure \ref{fig:param_trainrate} displays the results on three datasets. 

We observe from Figure \ref{fig:param_trainrate} that the AUC \HL{values increase} smoothly when the training ratio goes larger, demonstrating that more training data provides a better supervision signal for training. We can also find that even if the training data is rate ($20\%$), our framework still has a competitive performance, especially on two Bitcoin datasets. This observation shows that TADDY can learn an informative representation even trained with scarce data. Moreover, the variance of AUC decreases with the increase of the training ratio, which illustrates that our proposed framework tends to have a stable performance when training data is adequate.

\subsection{Ablation Study}

\begin{table}[!t]
	\small
	\centering
	\caption{Ablation study for TADDY and its variants on three datasets.}
	\begin{tabular}{ p{80 pt}|p{40 pt}<{\centering}p{40 pt}<{\centering}p{40 pt}<{\centering}}    
		\toprule[1.0pt]
		  & {UCI Messages} & {Bitcoin-Alpha} & {Bitcoin-OTC} \\
		\cmidrule{1-4}
		TADDY                        & 0.8370 & \bf{0.9423} & \bf{0.9262}\\		
		\texttt{w/o diff. enc.}               & 0.8304 & 0.9329 & 0.9153\\
		\texttt{w/o dist. enc.}               & 0.5362 & 0.5043 & 0.5187\\
        \texttt{w/o temp. enc.}               & \bf{0.8399} & 0.9326 & 0.9021\\
		\bottomrule[1.0pt]
	\end{tabular}
	\label{table:ablation}
\end{table}

To study the contribution of each component in the spatial-temporal node encoding towards the overall performance, we conduct the ablation study of the proposed TADDY framework. In particular, We evaluate the following variants of the node encoding: \texttt{w/o diff. enc.}, \texttt{w/o dist. enc.} and \texttt{w/o temp. enc.}, where the diffusion-based spatial encoding, distance-based spatial encoding and relative temporal encoding are discarded respectively when excusing the encoding fusion. We perform the ablation study with $10\%$ anomaly proportion for each dataset, and all the parameters are set as default. Table \ref{table:ablation} reports the proposed framework and its variants on three datasets. We have the following observations according to the results:

\begin{itemize} 

	\item The distance-based spatial encoding is the most critical term in node encoding. Without this term, the AUC \HL{values} decrease sharply to about $50\%$, which indicates that the anomalies become indistinguishable. This observation proves that the local structural information is significant in detecting anomalies, which is also pointed out in previous works \cite{cai2020structural,liu2021anomaly}. 
	
	\item The diffusion-based spatial encoding and relative temporal encoding both have a minor contribution in detecting anomalies. In the vast majority of cases, removing one of them would lead to a slight performance drop. We infer from the results that the diffusion-based spatial encoding provides a global view for graph structure which has a minor relation to anomaly detection. Moreover, the temporal encoding points out the occurrence time of neighborhoods which is relatively unimportant since the nodes have been selected to the substructure set.
	
	\item In most of the cases, combining all of the three types of encoding has the highest AUC \HL{values}, excepted on UCI Messages dataset. This shows that using a comprehensive spatial-temporal encoding is meaningful to anomaly detection. As for the exception, we guess that emphasizing the relative time distance may lead to an over-fitting on such \HL{a} property in some cases, which further results in a slight side effect on performance.
	
\end{itemize}

\section{Conclusion} \label{sec:conclusion}

In this paper, we make the first attempt to utilize transformer models for the graph anomaly detection problem \HL{in} dynamic graph \HL{scenarios}. We propose an end-to-end anomaly detection framework, TADDY, which is composed of four components: edge-based substructure sampling, spatial-temporal node encoding, dynamic graph transformer, and discriminative anomaly detector. Our framework constructs an informative and comprehensive node encoding to better represent the roles of nodes in an evolving graph space and successfully captures the coupled spatial-temporal information within dynamic graphs with a sole transformer model. Experiments on several real-world datasets show that the proposed framework detects anomalies with high effectiveness in dynamic graphs and outperforms the existing methods significantly. 


\section*{Acknowledgement}
This research was supported in part by  the Australian Research Council (ARC) under a Future Fellowship No. FT210100097 and National Natural Science Foundation of China project 61963004.

\ifCLASSOPTIONcaptionsoff
  \newpage
\fi

\bibliography{reference}
\bibliographystyle{IEEEtran}

\vspace{-1.5cm}

\begin{IEEEbiography}[{\includegraphics[width=1in,height=1.25in,clip,keepaspectratio]{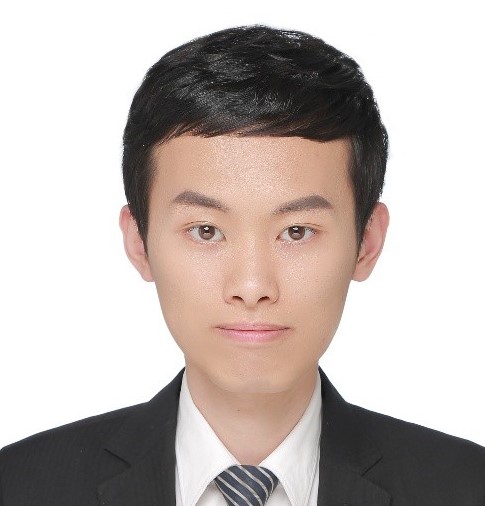}}]{Yixin Liu} received the B.S. degree and M.S. degree from Beihang University, Beijing, China, in 2017 and 2020, respectively. He is currently pursuing his Ph.D. degree in computer science at Monash University, Melbourne, Australia. His research concentrates on data mining, machine learning, and deep learning on graphs. 
\end{IEEEbiography}

\vspace{-1.5cm}

\begin{IEEEbiography}[{\includegraphics[width=1in,height=1.25in,clip,keepaspectratio]{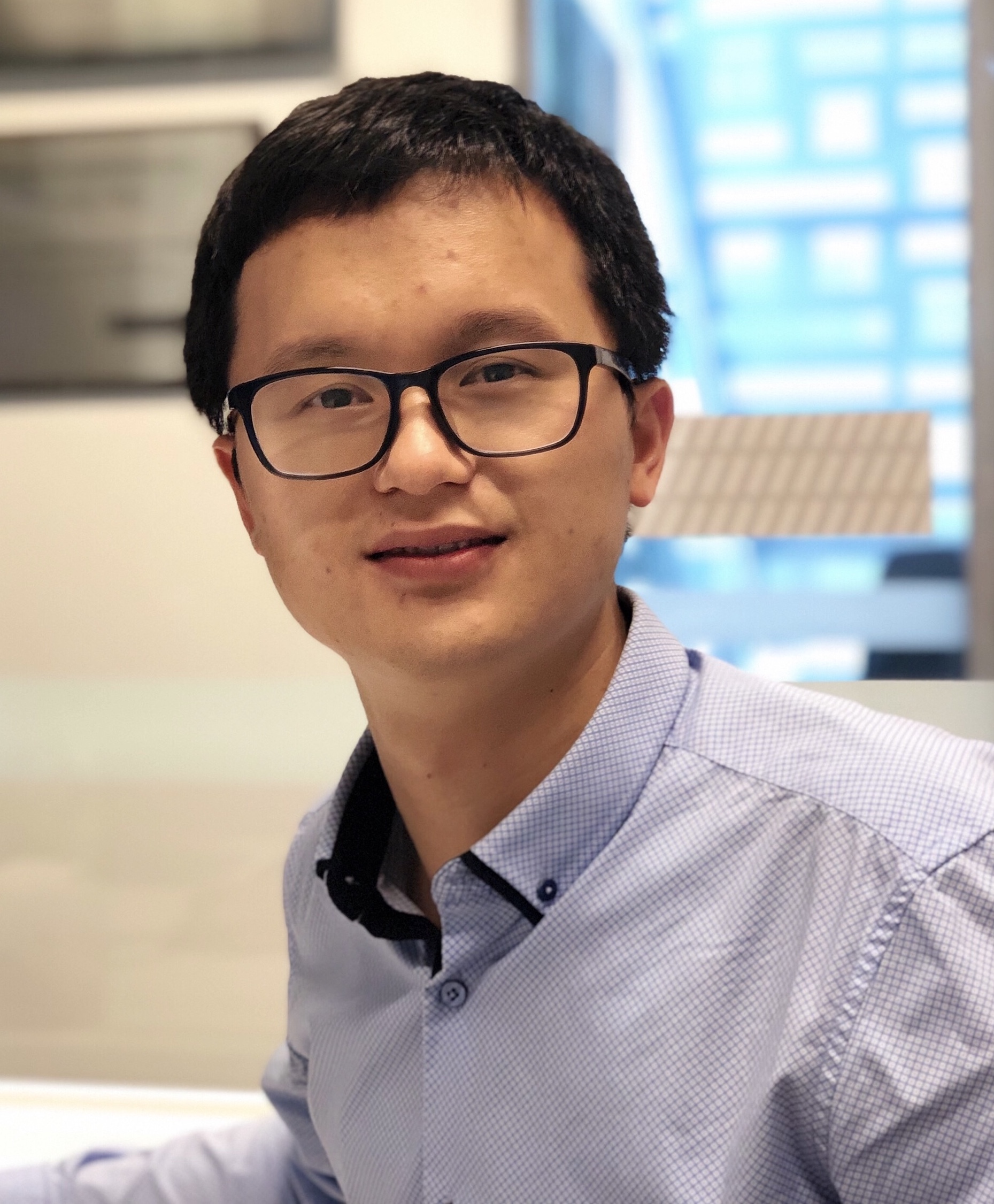}}]{Shirui Pan} received a Ph.D. in computer science from the University of Technology Sydney (UTS), Ultimo, NSW, Australia. He is currently a Senior Lecturer with the Faculty of Information Technology, Monash University, Australia. He is an
ARC Future Fellow (awarded in 2021). 
His research interests include data mining and machine learning. To date, Dr Pan has published over 100 research papers in top-tier journals and conferences, including TPAMI, TNNLS, and TKDE. 
\end{IEEEbiography}

\vspace{-1.5cm}

\begin{IEEEbiography}[{\includegraphics[width=1in,height=1.25in,clip,keepaspectratio]{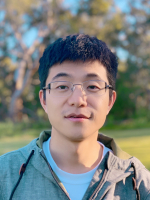}}]{Yu Guang Wang} received a Ph.D. in applied mathematics from University of New South Wales, Australia. He is an adjunct associate lecturer at UNSW Sydney. He is also a scientist at Max Planck Institute for Mathematics in Sciences, in Mathematics Machine Learning group. His research interests lie in computational mathematics, statistics, machine learning, and data science. 
\end{IEEEbiography}

\vspace{-1.5cm}

\begin{IEEEbiography}[{\includegraphics[width=1in,height=1.25in,clip,keepaspectratio]{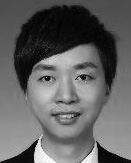}}]{Fei Xiong} received the Ph.D. degree from Beijing Jiaotong University, Beijing, China, in 2013.
He was a Visiting Scholar with Carnegie Mellon University, Pittsburgh, PA, USA, from 2011 to 2012. He is currently an Associate Professor with the School of Electronic and Information Engineering, Beijing Jiaotong University. His current research interests include Web mining, complex networks, and complex systems.
\end{IEEEbiography}

\vspace{-1.5cm}

\begin{IEEEbiography}[{\includegraphics[width=1in,height=1.25in,clip,keepaspectratio]{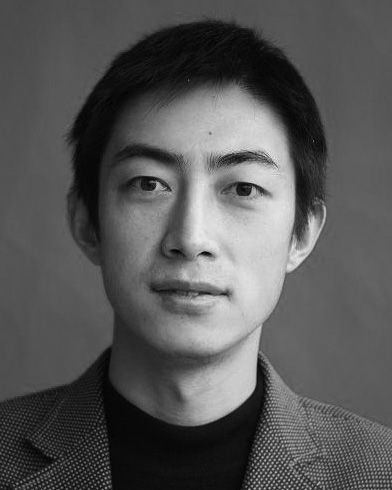}}]{Liang Wang}  the Ph.D. degree in computer science from the Shenyang Institute of Automation, Chinese Academy of Sciences, Shenyang, China, in 2014.
He was a Post-Doctoral Researcher with Northwestern Polytechnical University, Xi’an, China, in 2017, where he is currently an Associate Professor. His current research interests include ubiquitous computing, mobile crowd sensing, and data mining.
\end{IEEEbiography}

\vspace{-1.5cm}

\begin{IEEEbiography}[{\includegraphics[width=1in,height=1.25in,clip,keepaspectratio]{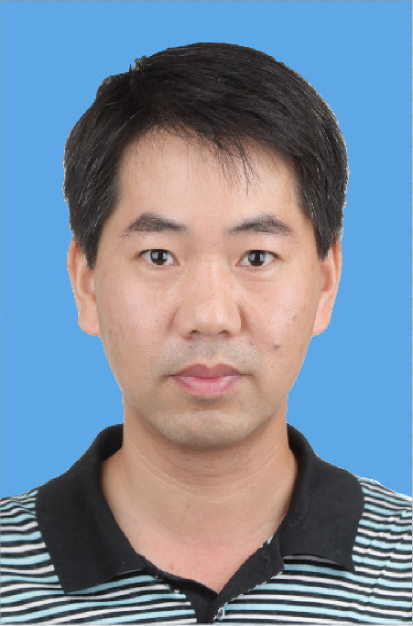}}]{Qingfeng Chen}  received the BSc and MSc degrees in mathematics from Guangxi Normal University, China, in 1995 and 1998, respectively, and the PhD degree in computer science from the University of Technology Sydney, in September 2004. He is now a professor with Guangxi University, China, and the Hundred Talent Program of Guangxi. His research interests include bioinformatics, data mining, and artificial intelligence. 
\end{IEEEbiography}

\vspace{-1.5cm}

\begin{IEEEbiography}[{\includegraphics[width=1in,height=1.25in,clip,keepaspectratio]{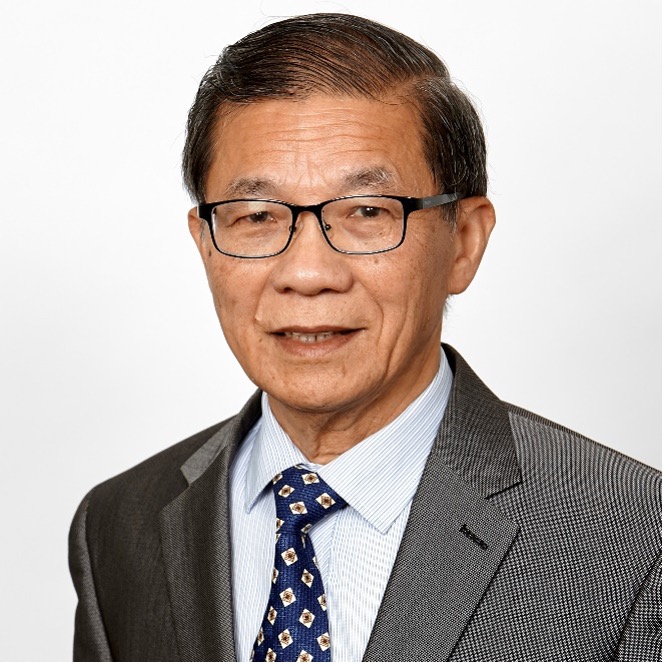}}]{Vincent CS Lee} received the PhD degree in adaptive systems from The University of NewCastle, Australia, in 1992. He is currently an associate Professor with the Department of Data Science and Artificial Intelligence, Faculty of IT, Monash University, Australia. He is a multi-interdisciplinary researcher spanning adaptive signal processing and control system, computational intelligence, AI in economic and finance, deep machine learning and computer vision, digital health process mining, and information security and network cryptography disciplines. He is a senior member of the IEEE.
\end{IEEEbiography}

\end{document}